\documentclass[11pt]{article}
\usepackage[a4paper,margin=1in]{geometry}
\usepackage[T1]{fontenc}
\usepackage[utf8]{inputenc}
\usepackage{lmodern,color}
\usepackage[nopatch=footnote]{microtype}
\usepackage{amsmath,amssymb,amsthm,mathtools,bm}
\usepackage{booktabs,tabularx,array}
\usepackage{graphicx}
\usepackage[section]{placeins}
\usepackage{enumitem}
\usepackage{algorithm}
\usepackage{algpseudocode}
\algrenewcommand{\algorithmicrequire}{\textbf{Input:}}
\algrenewcommand{\algorithmicensure}{\textbf{Output:}}
\algrenewcommand{\algorithmicindent}{1.2em}
\usepackage[authoryear,round]{natbib}
\usepackage[hidelinks]{hyperref}
 \hypersetup{pdftitle={Dependence-Aware Risk Assessment and Optimal Design of AI-Agent Safeguard Systems},
 pdfauthor={Maochao Xu},
 pdfsubject={Operations research; dependence-aware safeguard design},
 pdfkeywords={copula, risk assessment, layered screening, selective verification, dependence, safeguard design}}
\usepackage[nameinlink,noabbrev,capitalise]{cleveref}
\usepackage{setspace}
\newtheorem{theorem}{Theorem}[section]
\newtheorem{proposition}[theorem]{Proposition}

\newtheorem{corollary}[theorem]{Corollary}
\theoremstyle{remark}

\theoremstyle{plain}
\newtheorem{example}{Example}[section]
\numberwithin{equation}{section}
\DeclareMathOperator*{\argmin}{arg\,min}

\newcommand{\ind}{\mathbf 1}
\newcommand{\cA}{\mathcal A}
\newcommand{\cB}{\mathcal B}
\newcommand{\cQ}{\mathcal Q}
\newcommand{\bD}{\mathbf D}
\newcommand{\bw}{\mathbf w}
\newcommand{\sm}{\leq_{\mathrm{sm}}}
\crefname{algorithm}{Algorithm}{Algorithms}
\Crefname{algorithm}{Algorithm}{Algorithms}
\title{Dependence-Aware Risk Assessment and Optimal Design of AI-Agent Safeguard Systems}
   \author{Yijia Li$^1$ \quad Maochao Xu$^2$\thanks{Corresponding author. Email: mxu2@ilstu.edu}  \quad Peng Zhao$^3$\\ 
      $^1${\small School of Statistics and Data Science,}\\
{\small Shanghai University of International Business and Economics, China}
     \\
  $^2${\small Department of Mathematics, Illinois State University, USA}\\
 $^3${\small School of Mathematics and Statistics, Jiangsu Normal University,  China}\\
     }
\date{}

\begin{document}
\maketitle

\begin{abstract}
Large language model (LLM) agents increasingly rely on safeguards to prevent unauthorized actions while preserving legitimate service. However, existing risk assessments often focus on individual safeguard performance while overlooking dependence among safeguard decisions. This paper develops a dependence-aware framework for risk assessment and optimal design of multi-component AI safeguard systems, jointly selecting configurations and operating policies for sequential layered screening and selective verification. We establish stochastic comparisons of layered-screening error probabilities and invocation costs under lower-orthant ordering. For selective verification, we characterize exactly which policies have error probabilities and review costs that respond monotonically to stronger dependence under supermodular ordering. For a fixed first stage and reviewer, we show that the optimal routing map is a two-threshold rule in the first-stage likelihood ratio, and that a reviewer enters an optimal policy only when its cost is below a quality-adjusted combination of the error losses. We further develop exact optimization algorithms for selecting safeguard subsets, execution orders, and verification policies under specified risk and cost requirements. An accompanying open-source software package, Dependence-Aware Policy Optimization Algorithms (\href{https://github.com/MSMCI-lab/DAPOA}{DAPOA}), facilitates the practical implementation of the proposed framework. Numerical experiments show that ignoring dependence can lead to inaccurate risk assessments and infeasible safeguard designs, with independence-based selections violating security requirements in six of eight scenarios. An empirical study using four LLM safeguards on  benchmark-derived agent actions further demonstrates the practical implications of dependence modeling. Under the principal operating setting, dependence-aware selection lowers held-out total cost by 39.6\% relative to independence-based selection.
\end{abstract}

 \newpage

\section{Introduction}\label{sec:intro}

LLM agents increasingly act through tools that
retrieve information, modify records, execute code, and communicate with
external services. An agent can propose an action that is syntactically
valid but unauthorized, such as sending information to an unapproved
recipient or changing a record when the user requested only a read.
Such proposals can arise from ordinary model errors or from indirect
prompt injection, in which instructions embedded in external content
redirect the agent's behavior
\citep{greshake2023indirect,debenedetti2024agentdojo}. Developers therefore
deploy safeguards that screen agent inputs, outputs, and actions.
Examples include learned content classifiers such as Llama Guard
\citep{inan2023llamaguard}, programmable policy enforcement through NeMo
Guardrails \citep{rebedea2023nemo}, and the prompt-injection and
goal-alignment checks integrated in LlamaFirewall
\citep{chennabasappa2025llamafirewall}. When such components are combined, system risk depends on each safeguard’s performance, how their errors overlap, and the rule used to combine their decisions. A safeguard layer operates on demand and has two consequential failure
modes. A \emph{security miss} allows an unauthorized action, whereas a
\emph{false block} stops an authorized action and interrupts service. Current practice uses several approaches to combining safeguards. Two approaches that arise in practice are layered screening and cascade verification. In \emph{layered screening}, several safeguards are applied as successive layers to a proposed action, and the action proceeds only if every layer allows it. Thus, a blocking decision by any safeguard can stop the action. LlamaFirewall, for example, halts the agent when either its injection
detector or its goal-alignment check raises a flag
\citep{chennabasappa2025llamafirewall}. Under this rule, an additional layer
cannot increase the miss probability, but it can contribute false blocks of
its own. In LlamaFirewall's AgentDojo evaluation, task utility was 47.0\%
with the injection detector alone, 43.1\% with the alignment check alone, and
42.7\% with both \citep{chennabasappa2025llamafirewall}. For \emph{cascade verification} (i.e., flag-to-review), an inexpensive safeguard first screens the proposed action, and flagged actions are escalated to a more costly reviewer for further verification. Anthropic's Constitutional Classifiers++
provides a content-safety example: a lightweight linear probe screens all
exchanges and escalates suspicious ones to a second-stage classifier, whose
final decision combines the probe and classifier scores
\citep{cunningham2026classifiers}. A cascade blocks an action only when the
screen refers it and the reviewer confirms it, and it misses an attack
whenever either stage does. Its security-miss probability therefore depends on the screen's miss probability and on the reviewer's miss probability conditional on referral.

Current practice often evaluates and tunes safeguards largely in
isolation, reporting component detection and false-positive rates, or
system-level performance for a single fixed configuration. System risk under
alternative combination rules  is then either not assessed or
computed, implicitly, as if component errors were independent. This
assumption is difficult to justify in practice as safeguards in the same system inspect the
same context: an injected instruction that appears in a retrieved document is
presented to every layer that reads it. Many safeguards are themselves
language models, or probes on language-model activations, that share
pretraining corpora, architectures, and often a base model family; some are
trained on synthetic data generated from the same policy or constitution.
Attacks also vary in intrinsic difficulty, so an obfuscated or semantically
disguised instruction that defeats one detector tends to defeat others, a
phenomenon  recently
documented for the errors of LLMs \citep{kim2025correlated}.
  Finally,
architecture itself induces dependence: in a cascade, the reviewer sees only
the actions the screen selects, and its decision may reuse the screen's score
\citep{cunningham2026classifiers}. Ignoring this dependence biases the assessment of system risk.

Dependence modeling plays a central role in reliability
engineering and operations research. For example,
\citet{wu2024warranty} models the dependence between the age and
usage of items under two-dimensional warranty with copulas and
derives preventive maintenance policies from the resulting failure
process. \citet{rahimi2026inventory} characterize ordering policies
for multi-item inventory systems with autocorrelated and
cross-correlated demands and report substantial cost reductions
relative to policies that treat demands as independent. In
portfolio credit derivatives, the dependence among default times
drives pricing, and \citet{lei2023copula} estimate copula
sensitivities that quantify how changes in the dependence level
affect the output. \citet{kim2022robustrisk} use copula-based
models to assess worst-case average value at risk under tail
uncertainty, with an application to cyber risk. Cyber-risk research
has also quantified the losses caused by security failures:
\citet{eling2019cybercosts} analyze actual cyber-event costs using
operational risk data, and \citet{dacorogna2023cyber} develop a
method for modeling heavy-tailed cyber losses. For security
planning, \citet{paul2021cyberplanning} develop a two-stage
stochastic programming model for allocating prevention, detection,
containment, and intelligence resources, and \citet{li2026cloud}
represent dependent compromises in multi-tenant cloud systems and
connect joint risk assessment to defensive decisions. Where these
studies model dependence, it is among inputs, losses, or
compromised assets. In a safeguard system, the relevant dependence
is among the decisions of components that screen the same action.

Safeguard execution is also related to sequential testing, where
costly tests are ordered to determine a system's state.
\citet{rostami2019sequential} develop exact methods for testing
serial systems under precedence constraints, and
\citet{rostami2026sparse} study exact sequencing in sparse
precedence networks. \citet{agnetis2022timecritical} consider
testing and search when limited resources and deadlines restrict
the available decisions. In the safeguard setting studied here,
the decisions include the component subset, execution order,
reviewer, and final response to each first-stage pattern.
The objective accounts for invocation costs and the consequences
of both security misses and false blocks, using separate joint
decision distributions for unauthorized and authorized actions.
Stochastic comparisons provide a complementary way to assess
system designs under structural assumptions on dependence.
\citet{navarro2021redundancy} use copulas and distortion functions
to compare redundancy mechanisms and determine where redundant
components should be placed.   \citet{argyris2025dominance}
derive conditions for multidimensional stochastic dominance between
discrete joint distributions with respect to specified classes of
value functions. 



Correlated errors among LLMs have been documented
\citep{kim2025correlated}, and recent work shows that layered LLM
defenses can be defeated jointly by staged attacks
\citep{mckenzie2026stack}. To our knowledge, however, the joint
distribution of safeguard decisions has not been modeled and used
to compare alternative combination policies or to design how
safeguards are selected, ordered, and combined. This gap matters because the effect of the
joint distribution on system risk and cost depends on the policy
that combines the decisions. In layered screening, the joint
outcomes of the selected safeguards determine system errors, while
their execution order determines how often each safeguard is
invoked. In a cascade, the referral rule determines when the
reviewer is used and which actions rely on its decision. These
considerations raise two related questions: (i) how can joint
safeguard decisions be modeled and translated into policy-specific
measures of risk and cost; and (ii) how can these measures be used
to select safeguards, execution orders, reviewers, and referral
rules under operational constraints? To address these questions,
we introduce two policy classes for the design of AI-agent action
safeguards, building on layered screening and cascade verification
used in practice. \emph{Sequential layered
screening} selects and orders a subset of safeguards, stopping when
any selected safeguard returns BLOCK. \emph{Selective verification}
selects first-stage safeguards and a reviewer, then assigns each
joint first-stage decision pattern to automatic allowance,
automatic blocking, or review. These classes include fixed layered
configurations and flag-to-review cascades as special cases. We
assess and optimize their risk and operating cost using the
class-conditional joint distribution of safeguard decisions.

Our main contributions are as follows.
\begin{itemize}[leftmargin=1.6em]

\item \emph{Dependence-aware assessment and design formulation.}
We develop a copula-based framework for modeling the
class-conditional joint distribution of safeguard decisions.
For sequential layered screening and selective verification,
we derive system-level miss and false-block probabilities and
establish stochastic comparisons that characterize how dependence
among safeguard decisions affects policy risk and invocation
costs. For selective verification, we prove that a policy's
error probabilities respond monotonically to stronger dependence
if and only if its final-decision rule is supermodular or
submodular, and its referrals and review cost do so if and only if
its referral rule is. 


\item \emph{Dependence-Aware Policy Optimization Algorithms (DAPOA).}
We formulate two design problems: minimizing expected total
cost subject to risk requirements, and minimizing the
security-miss probability subject to a cost constraint. For a
fixed first stage and reviewer, we show that the optimal routing
map is a two-threshold rule in the first-stage likelihood ratio. DAPOA
uses subset dynamic programming to select safeguards and their
execution order for sequential layered screening, and binary
optimization to select safeguards, reviewers, referral regions,
and final decisions for selective verification. The searches return
an optimal feasible policy within the admitted classes for the
supplied class-conditional distributions. 
The algorithms are implemented in the open-source
(\href{https://github.com/MSMCI-lab/DAPOA}{DAPOA})\footnote{\url{https://github.com/MSMCI-lab/DAPOA}} package.

\item \emph{Numerical and empirical evaluation.}
Through numerical experiments, we identify when dependence
materially changes estimated system risk, policy feasibility,
and the optimal design. Using agent-action safeguard data, we
evaluate the fitted joint model and compare the risk and
operating costs of layered, cascade, and selective verification
policies. The results show how joint safeguard outcomes can be
translated into safeguard selection, execution order, and review
decisions, and delineate the conditions under which accounting
for dependence improves the risk--cost tradeoff.

\end{itemize}

The remainder of the paper is organized as follows.
Section~\ref{sec:system} defines the safeguard policies, introduces
the class-conditional copula representation, and formulates the
risk-constrained design problems. Section~\ref{sec:switching}
establishes the dependence comparisons, characterizes the verification
policies whose risk responds monotonically to dependence, derives the
structure of optimal routing maps, and develops exact algorithms
for safeguard selection, execution ordering, and selective verification.
Sections~\ref{sec:numerical} and~\ref{sec:empirical} present the numerical
and empirical studies, respectively. Section~\ref{sec:discussion}
discusses the findings, limitations, and future research directions.
The Supplementary Material provides copula definitions, family-specific
calculations,   and supplementary empirical methods
and results.

Table~\ref{tab:notation} summarizes the main notation used throughout
the paper.

\begin{table}[htb!]
\centering
\caption{Main notation.}
\label{tab:notation}
\footnotesize
\renewcommand{\arraystretch}{1}
\begin{tabular}{@{}p{0.25\textwidth}p{0.70\textwidth}@{}}
\toprule
Notation & Definition \\
\midrule
$Y$ &
True authorization status: $Y=1$ for an unauthorized action and
$Y=0$ for an authorized action \\

$m,\ j$ &
Number of available safeguards and safeguard index \\

$D_j,\ \mathbf{D},\ \mathbf{d}$ &
Decision of safeguard $j$; potential decision vector
$\mathbf{D}=(D_1,\ldots,D_m)$; and a realization
$\mathbf{d}\in\{0,1\}^m$ \\

$P_y,\ p_{jy},\ q_y(\mathbf{d})$ &
Class-conditional probability $P_y(\cdot)=\Pr(\cdot\mid Y=y)$;
marginal blocking probability $p_{jy}=P_y(D_j=1)$; and joint
pattern probability $q_y(\mathbf{d})=P_y(\mathbf{D}=\mathbf{d})$ \\

$\pi,\ G_\pi,\ \mathcal P$ &
Operating policy; final system decision; and admissible policy
class $\mathcal P=\mathcal P_L\cup\mathcal P_R$ \\

$M(\pi),\ B(\pi)$ &
Security-miss probability $P_1(G_\pi=0)$ and false-block
probability $P_0(G_\pi=1)$ \\

$c_j,\ \omega$ &
Cost of invoking safeguard $j$ and target proportion of
unauthorized actions \\

$K(\pi;\omega),\ J(\pi;\omega)$ &
Expected operating cost and total expected cost, respectively \\

$\ell_M,\ \ell_B$ &
Expected losses per security miss and false block \\

$\alpha,\ \beta,\ b$ &
Upper bounds on miss and false-block probabilities and operating-cost
budget \\

$S,\ \sigma$ &
Selected safeguard set and its execution order under layered screening \\

$F,\ h,\ H,\ W$ &
First-stage safeguard set; reviewer index; reviewer decision
$H=D_h$; and first-stage pattern $W=(D_j:j\in F)$ \\

$\mathcal A_\pi,\ \mathcal B_\pi,\ \mathcal Q_\pi$ &
Automatic-allow, automatic-block, and review regions under
selective verification \\

$s_y(\bw),\ t_y(\bw)$ &
First-stage pattern probability $P_y(W=\bw)$ and joint probability
$P_y(W=\bw,H=1)$ that the pattern occurs and the reviewer blocks \\

$\lambda(\bw),\ \kappa_y$ &
First-stage likelihood ratio   and the
reviewer's conditional BLOCK rate   \\
\bottomrule
\end{tabular}
\end{table}
\section{A Dependence-Aware Framework for Safeguard Risk Assessment and Design}
\label{sec:system}
This section first defines the system error measures and the two
classes of safeguard policies. Then, we introduce how the copula can be used to model joint distributions of
safeguard decisions. Finally, we define the cost criteria and
formulate the corresponding policy-design problems.

\subsection{System error measures and safeguard policies}
\label{subsec:policies}

Consider a candidate action proposed by an LLM agent. Let $Y\in\{0,1\}$ denote its true
authorization status, where $Y=1$ indicates that the action violates
the trusted authorization specification and $Y=0$ indicates that it is
authorized. Suppose that $m\geq 2$ safeguard components are available,
each returning a binary decision
\[
D_j=
\begin{cases}
0, & \text{if safeguard $j$ returns \textsc{allow}},\\
1, & \text{if safeguard $j$ returns \textsc{block}}.
\end{cases}
\]
The vector $\bD=(D_1,\ldots,D_m)$ collects the decisions that the
safeguards would return if all were applied to the same candidate under
a fixed evaluation protocol. A \emph{policy} $\pi$ specifies which components to invoke and how the observed
decisions determine the final system decision
$G_\pi\in\{0,1\}$, where $G_\pi=1$ denotes \textsc{block} and $G_\pi=0$ represents \textsc{allow}. The policy is
fixed before a new candidate is evaluated and cannot observe the true
label $Y$ or the decisions of safeguards that have not been invoked.
Writing
\[
P_y(\cdot)=\Pr(\cdot\mid Y=y),\qquad y\in\{0,1\},
\]
we define the security miss and false-block probabilities as
\begin{equation*}
 M(\pi)=P_1(G_\pi=0),\qquad
 B(\pi)=P_0(G_\pi=1).
\end{equation*}
The first is the probability of allowing an unauthorized action; the
second is the probability of blocking an authorized action.

In the following, we describe the proposed two policy classes.
\begin{itemize}
    \item {\em Sequential layered screening.}
A policy selects a nonempty subset
$S\subseteq\{1,\ldots,m\}$ of safeguards and an
invocation order $\sigma=(j_1,\ldots,j_{|S|})$.
The action is blocked as soon as any selected safeguard
returns BLOCK and is allowed only if all selected
safeguards return ALLOW.  Thus, the final decision is
\begin{equation*}
G_{\pi}
=\ind\left\{\sum_{j\in S}D_j\geq1\right\}.
\end{equation*}
The security miss and false-block probabilities are
\begin{equation*}
\begin{aligned}
M(\pi) =P_1(D_j=0,\ \forall j\in S),\quad
B(\pi) =1-P_0(D_j=0,\ \forall j\in S).
\end{aligned}
\end{equation*}
The selected set determines the final decision and its error
probabilities. The invocation order determines which safeguards are
reached before termination and therefore affects operating cost, which
is defined in Section~\ref{subsec:cost}. Fixing $S$ and $\sigma$ recovers a fixed layered screening configuration in practice.

\item  {\em Selective verification.} A two-stage policy selects a nonempty first-stage set
$F\subsetneq\{1,\ldots,m\}$ and a reviewer $H=D_h$, where $h\notin F$.
All safeguards in $F$ are invoked, producing the binary decision
pattern $W=(D_j:j\in F)$ in a fixed order. The pattern space
$\{0,1\}^{|F|}$ is partitioned into an
automatic-allow region $\cA_\pi$, an automatic-block region $\cB_\pi$,
and a review region $\cQ_\pi$. The reviewer is invoked only when
$W\in\cQ_\pi$, in which case its decision is final:
\begin{equation}\label{eq:selectiveg}
 G_\pi=\ind\{W\in\cB_\pi\}+\ind\{W\in\cQ_\pi\}H.
\end{equation}
The corresponding error probabilities are
\begin{align}
 M(\pi)&=P_1(W\in\cA_\pi)
          +P_1(W\in\cQ_\pi,H=0),\label{eq:selectM}\\
 B(\pi)&=P_0(W\in\cB_\pi)
          +P_0(W\in\cQ_\pi,H=1).\label{eq:selectB}
\end{align}
The design variables are $F$, $h$, and the routing partition.

The  \emph{flag-to-review} cascade is a special case of the
selective verification. It automatically allows the
unanimous-ALLOW pattern, refers every pattern containing at least one
BLOCK to the reviewer, and does not automatically block any pattern.  The final decision is
\begin{equation}\label{eq:cascade}
G_{\pi_{\rm F}}
=\ind\left\{\sum_{j\in F}D_j \geq1\right\}H.
\end{equation}
Its security miss and false-block probabilities are
\begin{align*}
M(\pi_{\rm F})
 &=P_1\left(\sum_{j\in F}D_j=0\right)+P_1\left(\sum_{j\in F}D_j \geq1,H=0\right),\\
B(\pi_{\rm F})
 &=P_0\left(\sum_{j\in F}D_j \geq1,H=1\right).
\end{align*}
 We retain the  flag-to-review cascade as a
practical baseline and as the policy used for the analytical cascade
comparison. The proposed framework,
however, searches the broader selective verification class, which can
treat different flagged patterns differently and can also review the
unanimous-ALLOW pattern.
\end{itemize}

\subsection{Copula modeling of joint safeguard decisions}
\label{subsec:copula}
Because the dependence among safeguard misses on unauthorized actions may differ from that among false blocks on authorized actions, we propose to model the joint safeguard decisions separately for the two classes.

Specifically, let $u_{jy}$ denote the conditional
marginal distribution function of $D_j$, for $j=1,\ldots,m$ and
$y\in\{0,1\}$. A copula $C_y$ provides a representation of the
class-conditional joint distribution:
\begin{equation*}
 P_y(\bD\leq\mathbf d)
 =C_y\!\left(u_{1y}(d_1),\ldots,u_{my}(d_m)\right).
\end{equation*}
 In particular, the probability
of a binary decision pattern $\mathbf d\in\{0,1\}^m$ is obtained by
finite differences:
\begin{equation}\label{eq:rect}
 q_y(\mathbf d)=P_y(\bD=\mathbf d)
 =\sum_{\boldsymbol\epsilon\in\{0,1\}^m}
 (-1)^{|\boldsymbol\epsilon|}
 C_y\!\left(u_{1y}(d_1-\epsilon_1),\ldots,
 u_{my}(d_m-\epsilon_m)\right),
\end{equation}
where $|\boldsymbol\epsilon|=\sum_j\epsilon_j$,
$u_{jy}(-1)=0$, $u_{jy}(0)=1-p_{jy}$, and
$u_{jy}(1)=1$ \citep{joe2014dependence}.
The  copula families provide
flexible representations of the pattern probabilities $q_0$ and
$q_1$, which determine the risk and cost measures used in
policy design.

For the policy classes discussed in this work, we present the explicit representations below.
\begin{itemize}
    \item For the sequential
layered screening, let $S=\{j_1,\ldots,j_s\}$, where $s=|S|$.
Let $p_{jy}=P_y(D_j=1)$, the security miss
and false-block probabilities can be represented as
\begin{equation*}
\begin{aligned}
M(\pi)
 =C_{1}(1-p_{j_1,1},\ldots,1-p_{j_s,1}), \quad 
B(\pi)
 =1-C_{0}(1-p_{j_1,0},\ldots,1-p_{j_s,0}).
\end{aligned}
\end{equation*}
Under the independence, these expressions reduce to
\begin{equation*}
 \begin{split}
 M^{\Pi}(\pi) =\prod_{j\in S}(1-p_{j1}),\quad
 B^{\Pi}(\pi) =1-\prod_{j\in S}(1-p_{j0}).
 \end{split}
\end{equation*}
\item For selective verification, define, for
$\mathbf w\in\{0,1\}^{|F|}$,
\begin{equation}\label{eq:theoryroutingprob}
\begin{aligned}
s_y(\mathbf w)&=P_y(W=\mathbf w)
=\sum_{\substack{\mathbf d\in\{0,1\}^m:\\ (d_j)_{j\in F}=\mathbf w}}q_y(\mathbf d),\\
t_y(\mathbf w)&=P_y(W=\mathbf w,H=1)
=\sum_{\substack{\mathbf d\in\{0,1\}^m:\\ (d_j)_{j\in F}=\mathbf w,\ d_h=1}}q_y(\mathbf d),
\end{aligned}
\end{equation}
the probability of the first-stage pattern and the joint
probability that the pattern occurs and the reviewer returns BLOCK.
The referral probability is
$P_y(W\in\cQ_\pi)=\sum_{\mathbf w\in\cQ_\pi}s_y(\mathbf w)$; for
the flag-to-review cascade with $F=\{j_1,\ldots,j_f\}$ it equals
\begin{equation*}
P_y(W\neq\mathbf 0)
=1-s_y(\mathbf 0)
=1-C_{y}(1-p_{j_1,y},\ldots,1-p_{j_f,y}).
\end{equation*}
The error probabilities in
Eqs.~\eqref{eq:selectM}--\eqref{eq:selectB} can be written as
\begin{align*}
M(\pi)
=\sum_{\mathbf w\in\cA_\pi}s_1(\mathbf w)
  +\sum_{\mathbf w\in\cQ_\pi}\{s_1(\mathbf w)-t_1(\mathbf w)\},
\quad
B(\pi)
=\sum_{\mathbf w\in\cB_\pi}s_0(\mathbf w)
  +\sum_{\mathbf w\in\cQ_\pi}t_0(\mathbf w).
\end{align*}
Clearly, dependence can affect not only
assessed risk but also the preferred component configuration and
routing policy.
\end{itemize}

In this work, we consider four popular copula families: Gaussian, Student-$t$, Clayton, and Gumbel copulas
\citep{joe2014dependence}. The Gaussian and Student-$t$ families
permit heterogeneous positive and negative association, with
the latter also allowing symmetric tail dependence. Clayton
and Gumbel capture positive dependence with lower- and
upper-tail asymmetry, respectively. The copula definitions
and parameter specifications are provided in the Supplementary Material.

\subsection{Costs and safeguard design}
\label{subsec:cost}

Safeguard design must account for security misses, false blocks, and the cost of invoking components. A policy determines both its error probabilities and which safeguards are invoked on a given action; the latter may vary because of early stopping or referral. We first define expected operating cost and the costs of incorrect decisions, then formulate two constrained design problems.

Let $c_j\geq0$ denote the operating cost incurred each time safeguard
$j$ is invoked, measured in monetary or appropriately normalized
computational units. For a potential decision vector $\mathbf d$, let
$\mathcal S_\pi(\mathbf d)$ denote the set of safeguards that policy
$\pi$ would invoke. The set actually invoked on a candidate is thus the
random set $\mathcal S_\pi(\bD)$. Because invocation can depend on
earlier decisions, this set may vary across candidates. The expected
operating cost per candidate is
\begin{equation*}
 K(\pi;\omega)=\sum_{j=1}^m c_j
 \left[\omega P_1\{j\in\mathcal S_\pi(\bD)\}
 +(1-\omega)P_0\{j\in\mathcal S_\pi(\bD)\}\right],
\end{equation*}
where $\omega\in[0,1]$ is the proportion of unauthorized actions in the
target deployment population.   Equivalently, in terms of the
pattern probabilities,
\begin{equation*}
 K(\pi;\omega)
 = \sum_{\mathbf d\in\{0,1\}^m}
 \left[\sum_{j\in\mathcal S_\pi(\mathbf d)}c_j\right]
 \left[\omega q_1(\mathbf d)+(1-\omega)q_0(\mathbf d)\right].
\end{equation*}
We treat each $c_j$ as fixed under the evaluation protocol. When token
usage varies across candidates, we use the mean cost as an
approximation.  We further consider costs of incorrect
system decisions. Let $\ell_M\geq0$ and $\ell_B\geq0$ denote the loss
per security miss and false block, respectively. The total expected
cost   is
\begin{equation}\label{eq:totalcost}
 J(\pi;\omega)=K(\pi;\omega)
 +\omega\ell_M M(\pi)
 +(1-\omega)\ell_B B(\pi).
\end{equation}
For the two policy classes, we have the following:
\begin{itemize}
    \item {Sequential layered screening.}
For an ordered policy $\sigma=(j_1,\ldots,j_{|S|})$, safeguard $j_r$ is
invoked only if all preceding safeguards return ALLOW. Hence,
\begin{equation*}
\begin{split}
K(\pi;\omega)
=\sum_{r=1}^{|S|}c_{j_r}\big[&\omega
P_1(D_{j_1}=\cdots=D_{j_{r-1}}=0)+(1-\omega)P_0(D_{j_1}=\cdots=D_{j_{r-1}}=0)\big],
\end{split}
\end{equation*}
where both probabilities equal one for $r=1$. Its total expected cost is
\begin{equation*}
J(\pi;\omega)
=K(\pi;\omega)
+\omega\ell_M M(S)+(1-\omega)\ell_B B(S).
\end{equation*}
\item {Selective verification.} For the first-stage set $F$ and reviewer $H$, the expected operating cost is
\begin{equation}\label{eq:reviewcost}
 K(\pi;\omega)=\sum_{j\in F}c_j+c_h\big[\omega P_1(W\in\cQ_\pi)
 +(1-\omega)P_0(W\in\cQ_\pi)\big].
\end{equation}
Its total expected cost follows from Eq.~\eqref{eq:totalcost}.   For the flag-to-review special case, Eq.~\eqref{eq:reviewcost} becomes
\begin{equation*}
K(\pi_{\rm F};\omega)
=\sum_{j\in F}c_j+c_h\big[
 \omega P_1(W\neq\mathbf 0)
 +(1-\omega)P_0(W\neq\mathbf 0)
\big].
\end{equation*}
\end{itemize}

\paragraph{Risk-constrained safeguard design.}

Let $\mathcal P=\mathcal P_L\cup\mathcal P_R$ be the finite set of admissible policies. A policy in $\mathcal P_L$ specifies a safeguard subset $S$ and an invocation order $\sigma$. A policy in $\mathcal P_R$ specifies a first-stage set $F$, a reviewer $h$, and a routing map that assigns each first-stage decision pattern to an automatic decision or to the reviewer, whose decision is then final. In practice, $\mathcal P$ can be constructed by retaining only policies that satisfy the application's operational requirements. These may restrict reviewer eligibility, limit the number of selected safeguards, or prescribe the routing action for particular patterns. When admissible, single-safeguard rules and fixed layered and flag-to-review configurations are included as comparison policies. The class specific copula modeling provides the probabilities needed to evaluate the risk and expected cost of each policy.
 
In practice, selecting a safeguard policy depends on how security, service disruption, and operating cost are prioritized. Minimizing total expected cost accounts for safeguard invocation and the consequences of both types of error, but the resulting policy may still exceed an acceptable miss or false-block rate. Our first formulation therefore imposes explicit limits on both errors. When preventing unauthorized actions is the primary goal, the second formulation minimizes the miss probability subject to an operating budget and a false-block limit. 

Specifically, the first objective is to minimize total expected cost subject to limits on both types of system error:
\begin{equation}\label{eq:designcost}
 \begin{aligned}
 \pi^{J}\in\operatorname*{arg\,min}_{\pi\in\mathcal P}
 &\quad J(\pi;\omega)\\
 \text{subject to}&\quad M(\pi)\leq\alpha,\qquad B(\pi)\leq\beta.
 \end{aligned}
\end{equation}
Here $\alpha$ and $\beta$ are prespecified upper bounds on the security-miss and false-block probabilities, respectively.
 The second problem minimizes the security-miss probability subject to an expected operating-cost budget and a false-block limit:
\begin{equation}\label{eq:designmiss}
 \begin{aligned}
 \pi^{M}\in\operatorname*{arg\,min}_{\pi\in\mathcal P}
 &\quad M(\pi)\\
 \text{subject to}&\quad K(\pi;\omega)\leq b,\qquad
 B(\pi)\leq\beta.
 \end{aligned}
\end{equation}
Here $b$ is an operating-cost budget. A problem is infeasible if no policy in $\mathcal P$ satisfies its constraints.

\section{Dependence Effects and Dependence-Aware Policy Optimization}
\label{sec:switching}

This section has three purposes. First, we establish how dependence
changes the risk and operating cost of fixed safeguard
architectures when component-level performance is held constant
(Sections~\ref{subsec:stochlayer} and~\ref{subsec:stochreview}).
For layered screening the direction of the change is always
determined; for selective verification we characterize exactly which
policies have a determined direction. Second, we
derive the structure of the optimal routing map for a fixed first
stage and reviewer (Section~\ref{subsec:exactdesign}): it is a
two-threshold rule in the first-stage likelihood ratio, and the
reviewer is used only when its cost is below an explicit function of
its conditional error rates and the error losses. Third, we present
the optimization algorithms that constitute DAPOA. These algorithms
incorporate copula-based modeling of dependence among safeguards to
select components and operating rules that optimize the specified
objectives subject to the stated constraints.

\subsection{Dependence effects in sequential layered screening}
\label{subsec:stochlayer}

For each authorization class $y\in\{0,1\}$,
consider two joint distributions $P_y^{(1)}$
and $P_y^{(2)}$ with identical marginals, and
let $\bD_y^{(r)}\sim P_y^{(r)}$, $r=1,2$.
We say that $\bD_y^{(2)}$ is smaller than
$\bD_y^{(1)}$ in the lower-orthant order if
\begin{equation}\label{eq:alloworder}
 \bD_y^{(2)}\leq_{\mathrm{lo}}\bD_y^{(1)}
 \quad\Longleftrightarrow\quad
 P_y^{(2)}(\bD\leq\mathbf d)
 \geq P_y^{(1)}(\bD\leq\mathbf d)
 \quad\text{for every }\mathbf d\in\mathbb R^m,
\end{equation}
where $\leq_{\mathrm{lo}}$ denotes the
lower-orthant order \citep{shaked2007stochastic}.
This ordering implies that any subset of
safeguards is at least as likely to issue
simultaneous ALLOW decisions under $P_y^{(2)}$
as under $P_y^{(1)}$. Related stochastic comparisons have been studied
for reliability systems with dependent components
\citep{navarro2021redundancy}.

For a fixed layered-screening policy $\pi$ with selected set $S$,
$s=|S|$, let $(j_1,\ldots,j_s)$ denote its ordered safeguards.
Under distribution $P_y^{(r)}$, let $N_\pi^{(r)}$
denote the random number of safeguards invoked.
The corresponding total invocation cost is
\[
V_\pi^{(r)}
=
\sum_{k=1}^{N_\pi^{(r)}}c_{j_k},
\]
where $c_j\geq0$ is the fixed cost of invoking
safeguard $j$.

\begin{proposition}\rm
\label{thm:layeredorder}
Suppose Eq.~\eqref{eq:alloworder} holds for
$y=0,1$. Then, for every layered-screening
policy $\pi$,
\[
M^{(1)}(\pi)\leq M^{(2)}(\pi),
\qquad
B^{(1)}(\pi)\geq B^{(2)}(\pi).
\]
Moreover, for each $y\in\{0,1\}$,
\[
N_\pi^{(1)}\mid Y=y
\leq_{\mathrm{st}}
N_\pi^{(2)}\mid Y=y,
\qquad
V_\pi^{(1)}\mid Y=y
\leq_{\mathrm{st}}
V_\pi^{(2)}\mid Y=y.
\]
Consequently,
\[
K^{(1)}(\pi;\omega)
\leq K^{(2)}(\pi;\omega),
\qquad \omega\in[0,1].
\]
\end{proposition}

The proof is given in Section~S2.1 of the Supplementary Material.
\Cref{thm:layeredorder} shows that dependence
among safeguard decisions affects both system reliability
and operating cost.
Under the specified lower-orthant ordering, a greater
probability of simultaneous ALLOW decisions leads to
a higher security-miss probability and expected
invocation cost, but a lower false-block probability.

 Table~\ref{tab:copulaorder} summarizes sufficient parameter
orderings under which the four copula families considered in
Section~\ref{subsec:copula} satisfy the lower-orthant ordering in
\Cref{thm:layeredorder}. The   full verifications are
provided in Section~S2.4 of the Supplementary Material; see also
\cite{joe2014dependence}.  

\begin{table}[htb!]
\centering
\caption{Sufficient within-family parameter conditions for the
lower-orthant ordering with fixed component margins.}
\label{tab:copulaorder}
\small
\setlength{\tabcolsep}{4pt}
\begin{tabularx}{\textwidth}{@{}lXc@{}}
\toprule
Copula family & Parameter ordering from model 1 to model 2
& Ordering~Eq.~\eqref{eq:alloworder} \\
\midrule
Gaussian
& $\Sigma_y^{(1)}\preceq_{\mathrm{off}}\Sigma_y^{(2)}$
& $\checkmark$ \\

Student-$t$
& $\Sigma_y^{(1)}\preceq_{\mathrm{off}}\Sigma_y^{(2)}$ and
  $\nu_y^{(1)}=\nu_y^{(2)}$
& $\checkmark$ \\

Clayton
& $0\leq\theta_y^{(1)}\leq\theta_y^{(2)}$
& $\checkmark$ \\

Gumbel
& $1\leq\theta_y^{(1)}\leq\theta_y^{(2)}$
& $\checkmark$ \\
\bottomrule
\end{tabularx}
\vspace{3pt}
\begin{minipage}{\textwidth}
\footnotesize
Here $y\in\{0,1\}$ denotes the authorization class and
$r\in\{1,2\}$ indexes the two dependence models.
$\Sigma_y^{(r)}$ is the positive-definite copula correlation
matrix, and $\Sigma_y^{(1)}\preceq_{\mathrm{off}}\Sigma_y^{(2)}$
means $[\Sigma_y^{(1)}]_{ij}\leq[\Sigma_y^{(2)}]_{ij}$ for every
$i\neq j$. The Student-$t$ parameter $\nu_y^{(r)}>0$ denotes
degrees of freedom; $\theta_y^{(r)}$ is the Clayton or Gumbel
dependence parameter.
\end{minipage}
\end{table}

\subsection{Dependence effects in selective verification}
\label{subsec:stochreview}

Unlike layered screening, selective verification assigns different
actions to different first-stage decision patterns. Its error
probabilities depend on the automatic-block and review regions
together with the reviewer's decision, while its operating cost
depends on the probability of referral.  We therefore use the supermodular order to relate
changes in dependence to the interaction structure of the
decision and referral rules.

For two distributions of $(W,H)$ on $\{0,1\}^{|F|+1}$, write
$P_y^{(1)}\leq_{\mathrm{sm}}P_y^{(2)}$ if
\[
E_y^{(1)}[a(W,H)]
\leq E_y^{(2)}[a(W,H)]
\]
for every supermodular function $a$
\citep{shaked2007stochastic}. A real-valued function on a binary
decision space is supermodular when every two-coordinate
cross-difference
\[
\Delta_{ij}a(\mathbf z)
=
a(0,0,\mathbf z)+a(1,1,\mathbf z)
-a(1,0,\mathbf z)-a(0,1,\mathbf z)
\]
is nonnegative, with the remaining coordinates fixed at
$\mathbf z$. It is submodular when all these differences are
nonpositive. 


\begin{theorem} \rm 
\label{thm:selectivesm}Fix a selective verification policy $\pi$.
Let $G_\pi$ be its final-decision rule in Eq.~\eqref{eq:selectiveg},  and let
$r_\pi(\bw)=\ind\{\bw\in\cQ_\pi\}$ be its referral indicator.
\begin{itemize}
\item[a)] The following are equivalent:
\begin{itemize}
\item[(i)] $G_\pi$ is supermodular (submodular);
\item[(ii)] $M^{(2)}(\pi)\leq M^{(1)}(\pi)$ ($\geq$) whenever
$P_1^{(1)}\sm P_1^{(2)}$;
\item[(iii)] $B^{(2)}(\pi)\geq B^{(1)}(\pi)$ ($\leq$) whenever
$P_0^{(1)}\sm P_0^{(2)}$.
\end{itemize}
\item[b)] The following are equivalent:
\begin{itemize}
\item[(i)] $r_\pi$ is supermodular (submodular);
\item[(ii)] $P_y^{(2)}(W\in\cQ_\pi)\geq P_y^{(1)}(W\in\cQ_\pi)$
($\leq$) whenever $P_y^{(1)}\sm P_y^{(2)}$, for $y=0,1$.
\end{itemize}
In this case, $K^{(2)}(\pi;\omega)\geq K^{(1)}(\pi;\omega)$
($\leq$) whenever $P_y^{(1)}\sm P_y^{(2)}$ for $y=0,1$, by
Eq.~\eqref{eq:reviewcost}.
\end{itemize}
\end{theorem}

The proof is given in Section~S2.2 of the Supplementary Material.
\Cref{thm:selectivesm}  shows
that in each case the sign structure of the rule is the sufficient and necessary 
condition: a policy's errors move in the same direction under
every supermodular increase in dependence if and only if its
final-decision rule is supermodular or submodular, and its
referrals do so if and only if its referral rule is. A supermodular
final-decision rule increases the BLOCK probability in each
authorization class, reducing security misses while increasing
false blocks; a submodular rule gives the opposite comparisons.
Operating cost follows the referral rule: it increases for a
supermodular referral indicator and decreases for a submodular
one. 

We present the following example for illustration.

\begin{example}\rm\label{ex:twosafeguards}
Assume there are two first-stage safeguards and a reviewer. Consider the following two scenarios:
\begin{itemize}
    \item Let the policy
refer only the pattern $11$ and automatically allow $00$, $10$,
and $01$. Then
\[
 \cA_\pi=\{00,10,01\},\qquad
 \cB_\pi=\varnothing,\qquad
 \cQ_\pi=\{11\},
\]
so $G_\pi(d_1,d_2,h)=d_1d_2h$ and
$r_\pi(d_1,d_2)=d_1d_2$ are both supermodular.
For $y=0,1$, let $P_y^{(1)}$ assign probability $1/8$ to each
configuration $(D_1,D_2,H)$. Obtain $P_y^{(2)}$ by adding
$\varepsilon=0.025$ to the unconditional cell probabilities of
$(0,0,1)$ and $(1,1,1)$ and subtracting it from $(1,0,1)$ and
$(0,1,1)$. Each component margin remains $1/2$, and
$P_y^{(1)}\leq_{\mathrm{sm}}P_y^{(2)}$.

The BLOCK probability increases from $0.125$ to $0.150$ and
the referral probability from $0.250$ to $0.275$ in each class.
Thus the changes have exactly the directions shown by
\Cref{thm:selectivesm}:
\[
 \begin{aligned}
 M^{(1)}(\pi)=0.875&\ \longrightarrow\ M^{(2)}(\pi)=0.850,\\
 B^{(1)}(\pi)=0.125&\ \longrightarrow\ B^{(2)}(\pi)=0.150,\\
 K^{(2)}(\pi;\omega)-K^{(1)}(\pi;\omega)&=0.025c_h>0,
 \qquad 0\leq\omega\leq1.
 \end{aligned}
\]
\item Let the policy refer any pattern other than $00$:
\[
 \cA_{\pi_{\rm F}}=\{00\},\qquad
 \cB_{\pi_{\rm F}}=\varnothing,\qquad
 \cQ_{\pi_{\rm F}}=\{10,01,11\}.
\]
The referral indicator
$r_{\pi_{\rm F}}(d_1,d_2)=d_1\vee d_2$ is submodular.
Thus \Cref{thm:selectivesm} shows fewer referrals
and lower operating cost under the same dependence change.
Indeed, the referral probability falls from $0.750$ to
$0.725$ in each class. 

Note that the   BLOCK-decision function
$G_{\pi_{\rm F}}(d_1,d_2,h)=h(d_1\vee d_2)$ has
$\Delta_{12}G_{\pi_{\rm F}}=-1$ at $H=1$ but
$\Delta_{1H}G_{\pi_{\rm F}}=1$ at $D_2=0$, so it is neither
supermodular nor submodular. By \Cref{thm:selectivesm}(a), the
direction of its error probabilities under supermodular increases
is therefore not determined; \Cref{ex:flagambiguity} below exhibits
both directions. 
\end{itemize}
\end{example}

For the flag-to-review cascade, we have the following result.
\begin{corollary}\rm\label{cor:flag}
Let $\pi_{\rm F}$ be the flag-to-review cascade in
Eq.~\eqref{eq:cascade}. Its referral indicator
$r_{\pi_{\rm F}}(\mathbf w)=1-\prod_{j\in F}(1-w_j)$ is submodular,
so under every classwise supermodular increase in dependence its
referral probabilities and operating cost cannot increase. If
$|F|\geq2$, its final-decision rule
$G_{\pi_{\rm F}}(\mathbf w,h)=h\,r_{\pi_{\rm F}}(\mathbf w)$ is neither
supermodular nor submodular, so by \Cref{thm:selectivesm}(a) the
direction of its miss and false-block probabilities is
undetermined: some supermodular increases raise them and others
lower them.
\end{corollary}


The following example exhibits the two opposite directions
explicitly for a cascade with two first-stage safeguards.

\begin{example}\normalfont\label{ex:flagambiguity}
Take $F=\{1,2\}$ and refer a pattern whenever at least one
first-stage safeguard returns BLOCK, allowing it otherwise. Thus
\[
G_{\pi_{\rm F}}(d_1,d_2,h)=h(d_1\vee d_2).
\]
This function is neither supermodular nor submodular:
$\Delta_{12}G_{\pi_{\rm F}}=-1$ when $h=1$, whereas
$\Delta_{1H}G_{\pi_{\rm F}}=1$ when $d_2=0$.
 
In each class, let the baseline assign probability $1/8$ to
every configuration $(D_1,D_2,H)$. Construct two alternatives,
each from this baseline, using a mass transfer of size
$\varepsilon=0.025$. The $D_1$--$D_2$ transfer adds
$\varepsilon$ to $(0,0,1)$ and $(1,1,1)$ and subtracts it
from $(1,0,1)$ and $(0,1,1)$. The $D_1$--$H$ transfer adds
$\varepsilon$ to $(0,0,0)$ and $(1,0,1)$ and subtracts it
from $(1,0,0)$ and $(0,0,1)$. Both preserve the component
margins and increase the class-conditional distribution in
supermodular order. Applying the same transfer in both classes
gives
\[
\begin{array}{lcccc}
 & P_y(W\in\cQ_{\pi_{\rm F}}) & M(\pi_{\rm F})
 & B(\pi_{\rm F}) & \Delta K\\
\hline
\text{Uniform baseline}       & 0.750 & 0.625 & 0.375 & 0\\
\text{$D_1$--$D_2$ transfer}  & 0.725 & 0.650 & 0.350 & -0.025c_h\\
\text{$D_1$--$H$ transfer}    & 0.750 & 0.600 & 0.400 & 0
\end{array}
\]
where $\Delta K$ is the change in operating cost from the
baseline and both first-stage safeguards are always invoked.
 
The first transfer reduces referrals and reviewer cost; the
second leaves both unchanged, consistent with
\Cref{cor:flag}. Their effects on misses and false blocks
reverse because the transfers act at cross-differences of
$G_{\pi_{\rm F}}$ with opposite signs. Thus supermodular
ordering alone gives no common direction for these errors
under this routing rule, as \Cref{thm:selectivesm}(a) implies
for any rule that is neither supermodular nor submodular.
\end{example}
 
\subsection{Dependence-Aware Policy Optimization Algorithms (DAPOA)}
\label{subsec:exactdesign}

This section develops DAPOA, a unified procedure for
optimizing safeguard policies. The algorithms select safeguards, execution orders, reviewers,
and routing decisions as applicable to each policy class.
An objective selector $O\in\{J,M\}$ determines which
design problem is solved: $O=J$ gives the
minimum-total-cost problem Eq.~\eqref{eq:designcost}, and
$O=M$ gives the minimum-miss problem
Eq.~\eqref{eq:designmiss}.

DAPOA searches the two policy classes separately and compares
their best feasible policies under the selected objective.
For sequential layered screening, it first finds the least-cost
execution order for every candidate safeguard subset and then
selects a feasible subset. For selective verification, it
evaluates candidate first-stage sets and reviewers and optimizes
the allow, block, or refer assignment of each first-stage
decision pattern. Both searches use the fitted $q_0$ and $q_1$ in Eq.~\eqref{eq:rect} to evaluate system errors and operating cost.
 
 \subsubsection{Selecting and ordering layered safeguards}

For a fixed subset $S$, layered screening allows an action exactly
when every safeguard in $S$ returns ALLOW, and, under any execution
order, a safeguard is invoked exactly when every safeguard before
it has returned ALLOW. The probability that a set of safeguards
jointly returns ALLOW therefore determines both the error
probabilities of $S$ and the expected cost of each execution order.
For $S\subseteq\{1,\ldots,m\}$, define the class-conditional   by
$$a_y(S)
  =\sum_{\mathbf d:\,d_j=0\ (j\in S)}q_y(\mathbf d),$$
and the deployment-mixture probability:
\begin{equation}\label{eq:theoryincrement}
\begin{aligned}
\bar a(S)
   =\omega a_1(S)+(1-\omega)a_0(S),
\end{aligned}
\end{equation}
where $a_y(\varnothing)=1$.  Let $K^*(S)$ be the minimum expected operating cost over all
execution orders of $S$. If safeguard $j$ is placed last, it is
invoked only when every safeguard in $S\setminus\{j\}$ returns
ALLOW. Its contribution to expected cost is therefore
$c_j\bar a(S\setminus\{j\})$. Note that the probability $\bar a(S\setminus\{j\})$  depends only on their
set, not on their order. Thus,  those
safeguards should be arranged at their minimum expected cost.
Minimizing over the choice of $j$ gives
\begin{equation}\label{eq:layeredcostdp}
\begin{aligned}
K^*(S)&=\min_{j\in S}
\left\{
K^*(S\setminus\{j\})
+c_j\bar a(S\setminus\{j\})
\right\},
\qquad S\ne\varnothing,
\end{aligned}
\end{equation}
Here $K^*(\varnothing)=0$. 
Therefore, a cost-minimizing order can be achieved by choosing a
minimizing $j$ in the recurrence Eq.~\eqref{eq:layeredcostdp} as the last safeguard and
repeating this step for $S\setminus\{j\}$.

The feasible selected sets for the two design objectives are
\begin{equation}\label{eq:layeredfeasible}
\begin{aligned}
\mathcal F_J
&=\{S\in\mathcal S_{\rm adm}:
      a_1(S)\leq\alpha,\quad 1-a_0(S)\leq\beta\},\\
\mathcal F_M
&=\{S\in\mathcal S_{\rm adm}:
      K^*(S)\leq b,\quad 1-a_0(S)\leq\beta\},
\end{aligned}
\end{equation}
where 
 $\mathcal S_{\rm adm}$ is the prespecified collection of
allowable nonempty safeguard subsets of $\{1,\ldots,m\}$.

For $O=J$, the terminal subset minimizes
\begin{equation*}
 K^*(S)+\omega\ell_Ma_1(S)
 +(1-\omega)\ell_B\{1-a_0(S)\}
 \quad\text{over }S\in\mathcal F_J.
\end{equation*}
For $O=M$, it minimizes $a_1(S)$ over
$S\in\mathcal F_M$. Algorithm~\ref{alg:layeredexact} solves both design problems
for sequential layered screening with early stopping.
The returned policy invokes safeguards in the recovered order
and stops at the first BLOCK.

Given $q_0$ and $q_1$, all subset probabilities
$a_y(S)$ can be computed  in
$O(m2^m)$ time. Algorithm~\ref{alg:layeredexact} then
evaluates $2^m$ states and $m2^{m-1}$ transitions.
Its time complexity is $O(m2^m)$. Although the running time of Algorithm~\ref{alg:layeredexact}
grows exponentially with $m$, exact selection and ordering
are practical.
Published safeguard systems include configurations with a modest
number of components, such as the two-scanner combination evaluated
for LlamaFirewall and the two-stage cascade used in Constitutional
Classifiers++ \citep{chennabasappa2025llamafirewall,
cunningham2026classifiers}. Exact optimization is practical for
candidate pools of this scale: with $m=5$, the layered recurrence
visits 32 subsets and evaluates 80 transitions.

\begin{algorithm}[htb!]
\caption{Optimal policy and objective value for sequential layered screening}
\label{alg:layeredexact}
\small\setstretch{1.0}
\begin{algorithmic}[1]
\Require $a_0,a_1$, $(c_j)_{j=1}^m$, $\omega$, $\beta$,
$O\in\{J,M\}$, $\mathcal S_{\rm adm}$;
$\alpha,\ell_M,\ell_B$ if $O=J$, and $b$ if $O=M$
\Ensure $(\pi_L^*,v_L^*)$; or \textsc{Infeasible}
\State $\bar a(S)\gets\omega a_1(S)+(1-\omega)a_0(S)$ for all $S$
\State $K^*(\varnothing)\gets0$
\For{$s=1,\ldots,m$}
  \ForAll{$S\subseteq\{1,\ldots,m\}$ with $|S|=s$}
    \State $j^*(S)\in\argmin_{j\in S}
      \{K^*(S\setminus\{j\})+c_j\bar a(S\setminus\{j\})\}$
    \State $K^*(S)\gets K^*(S\setminus\{j^*(S)\})
      +c_{j^*(S)}\bar a(S\setminus\{j^*(S)\})$
  \EndFor
\EndFor
\If{$O=J$}
  \State $\mathcal F\gets\mathcal F_J$ from Eq.~\eqref{eq:layeredfeasible}
\Else
  \State $\mathcal F\gets\mathcal F_M$ from Eq.~\eqref{eq:layeredfeasible}
\EndIf
\If{$\mathcal F=\varnothing$}
  \State \Return \textsc{Infeasible}
\EndIf
\If{$O=J$}
  \State $S^*\in\argmin_{S\in\mathcal F}
  \{K^*(S)+\omega\ell_Ma_1(S)+(1-\omega)\ell_B[1-a_0(S)]\}$
  \State $v_L^*\gets K^*(S^*)+\omega\ell_Ma_1(S^*)
                  +(1-\omega)\ell_B[1-a_0(S^*)]$
\Else
  \State $S^*\in\argmin_{S\in\mathcal F}a_1(S)$;
         $v_L^*\gets a_1(S^*)$
\EndIf
\State $T\gets S^*$; initialize $\sigma^*$ as the empty sequence
\While{$T\ne\varnothing$}
  \State $j\gets j^*(T)$; prepend $j$ to $\sigma^*$
  \State $T\gets T\setminus\{j\}$
\EndWhile
\State \Return $(\pi_L^*,v_L^*)$ with $\pi_L^*=(S^*,\sigma^*)$
\end{algorithmic}
\end{algorithm}

\subsubsection{Selecting the first stage, reviewer, and routing map}
To optimize selective verification, we consider each admissible
first-stage--reviewer pair $(F,h)$, with $h\notin F$, and optimize
its routing map. A routing map assigns each first-stage pattern
$\mathbf w\in\{0,1\}^{|F|}$ to automatic ALLOW,
automatic BLOCK, or REVIEW. For a referred pattern,
the reviewer's decision determines the final outcome. Every
safeguard in $F$ is invoked before routing; reviewer $h$ is invoked
only for a referred pattern. Write $\mathcal W_F=\{0,1\}^{|F|}$ for the pattern space, and
recall $s_y(\mathbf w)$ and $t_y(\mathbf w)$ from
Eq.~\eqref{eq:theoryroutingprob}.
 
For a routing partition $(\cA_\pi,\cB_\pi,\cQ_\pi)$ of $\mathcal W_F$,
we have, 
\begin{equation}\label{eq:routingcoeff}
\begin{aligned}
M(\pi)
 &=\sum_{\bw\in\cA_\pi}s_1(\bw)
   +\sum_{\bw\in\cQ_\pi}\{s_1(\bw)-t_1(\bw)\},\qquad
B(\pi)
 =\sum_{\bw\in\cB_\pi}s_0(\bw)
   +\sum_{\bw\in\cQ_\pi}t_0(\bw),\\
K(\pi;\omega)
 &=c_F+c_h\sum_{\bw\in\cQ_\pi}
       \{\omega s_1(\bw)+(1-\omega)s_0(\bw)\},
\end{aligned}
\end{equation}
where  
$c_F=\sum_{j\in F}c_j$.
For $O=J$, the routing map for a fixed $(F,h)$ solves
\begin{equation}\label{eq:routingMILP}
\min_{(\cA_\pi,\cB_\pi,\cQ_\pi)}\ J(\pi;\omega)
\quad\text{subject to}\quad
M(\pi)\leq\alpha,\qquad B(\pi)\leq\beta,
\end{equation}
where the minimum is over partitions of $\mathcal W_F$. For $O=M$,
it solves
\begin{equation}\label{eq:routingMILPmiss}
\min_{(\cA_\pi,\cB_\pi,\cQ_\pi)}\ M(\pi)
\quad\text{subject to}\quad
K(\pi;\omega)\leq b,\qquad B(\pi)\leq\beta.
\end{equation}
The quantities in Eq.~\eqref{eq:routingcoeff}, and hence
$J(\pi;\omega)$ in Eq.~\eqref{eq:totalcost}, are linear in the
indicators $\ind\{\bw\in\cB_\pi\}$ and $\ind\{\bw\in\cQ_\pi\}$, so
both problems are binary linear programs with two binary variables
per pattern. Operational restrictions can be imposed by fixing the
region of particular patterns.

\paragraph{Structure of the optimal routing map.}
The binary programs Eqs.~\eqref{eq:routingMILP}--\eqref{eq:routingMILPmiss}
return an optimal routing map for each pair $(F,h)$. It is interesting to study: which patterns are
referred to the reviewer and why, whether a reviewer is worth
including at all, and where dependence among the safeguards enters
the routing decision.   For a fixed pair $(F,h)$, let
\begin{equation}\label{eq:patterncost}
\begin{aligned}
\varphi_{\cA}(\bw)
 &=\omega\ell_M s_1(\bw),\quad
\varphi_{\cB}(\bw)
  =(1-\omega)\ell_B s_0(\bw),\\
\varphi_{\cQ}(\bw)
 &=c_h\{\omega s_1(\bw)+(1-\omega)s_0(\bw)\}\\
 &\quad+\omega\ell_M\{s_1(\bw)-t_1(\bw)\}
       +(1-\omega)\ell_B t_0(\bw)
\end{aligned}
\end{equation}
be the expected losses from automatically allowing, automatically
blocking, or referring pattern $\bw$, the last including the
reviewer's invocation cost. Substituting Eq.~\eqref{eq:routingcoeff}
into Eq.~\eqref{eq:totalcost} gives
\[
J(\pi;\omega)=c_F
+\sum_{\bw\in\cA_\pi}\varphi_{\cA}(\bw)
+\sum_{\bw\in\cB_\pi}\varphi_{\cB}(\bw)
+\sum_{\bw\in\cQ_\pi}\varphi_{\cQ}(\bw),
\]
so each pattern contributes only the loss of the action assigned
to it.

Consider first minimizing $J(\pi;\omega)$ over all
routing maps for $(F,h)$, that is, Eq.~\eqref{eq:routingMILP} without the
constraints. For a fixed $(F,h)$, total expected cost is $c_F$ plus the
sum of the pattern losses in Eq.~\eqref{eq:patterncost}. Thus, in
the absence of risk constraints, a routing map minimizes
$J(\pi;\omega)$ if and only if it assigns each pattern a
choice with the smallest pattern loss. In particular, REVIEW
is a minimum-loss choice at $\bw$ if and only if its loss is
no greater than either the automatic-ALLOW or automatic-BLOCK
loss. These two comparisons are equivalent to
\begin{equation}\label{eq:reviewpatternconditions}
\begin{aligned}
\omega\ell_M t_1(\bw)
&\geq c_h\{\omega s_1(\bw)+(1-\omega)s_0(\bw)\}
       +(1-\omega)\ell_B t_0(\bw),\\
(1-\omega)\ell_B\{s_0(\bw)-t_0(\bw)\}
&\geq c_h\{\omega s_1(\bw)+(1-\omega)s_0(\bw)\}
       +\omega\ell_M\{s_1(\bw)-t_1(\bw)\}.
\end{aligned}
\end{equation}
If both inequalities are strict, REVIEW is the unique
minimum-loss choice for $\bw$. Equality in either comparison
means that REVIEW ties with at least one automatic decision. The first inequality compares REVIEW with automatic ALLOW.
Reviewing avoids miss losses when the reviewer blocks an
unauthorized action, but incurs invocation cost and may
falsely block an authorized action. The second compares REVIEW
with automatic BLOCK. Reviewing avoids false-block losses
when the reviewer allows an authorized action, but incurs
invocation cost and may allow an unauthorized action. 
If an unconstrained cost-minimizing routing map also satisfies
$M(\pi)\leq\alpha$ and $B(\pi)\leq\beta$, it solves
Eq.~\eqref{eq:routingMILP}. The next result gives a
likelihood-ratio threshold rule. Let $\mathcal H_{\rm adm}$ denote the set of admissible
first-stage reviewer pairs:
\[
\mathcal H_{\rm adm}\subseteq
\left\{(F,h):
\varnothing\neq F\subseteq\{1,\ldots,m\},\quad
h\in\{1,\ldots,m\}\setminus F
\right\}.
\]
This set incorporates operational restrictions on which
safeguards may form the first stage and serve as the reviewer.

\begin{theorem}\rm
\label{thm:routingstructure}
Fix $(F,h)\in\mathcal H_{\rm adm}$. Suppose $s_0(\bw)>0$ and
$t_y(\bw)=\kappa_y s_y(\bw)$
for all $\bw$ and $y=0,1$, with constants
$\kappa_0,\kappa_1\in[0,1]$. Let $\lambda(\bw)=s_1(\bw)/s_0(\bw)$
and
\begin{equation}\label{eq:thresholds}
\begin{aligned}
\lambda_A&=\min\left\{
\frac{(1-\omega)\ell_B}{\omega\ell_M},\;
\frac{(1-\omega)(c_h+\ell_B\kappa_0)}
     {\omega(\ell_M\kappa_1-c_h)^+}
\right\},\\
\lambda_B&=\max\left\{
\frac{(1-\omega)\ell_B}{\omega\ell_M},\;
\frac{(1-\omega)\{\ell_B(1-\kappa_0)-c_h\}^+}
     {\omega\{c_h+\ell_M(1-\kappa_1)\}}
\right\},
\end{aligned}
\end{equation}
with $x/0=+\infty$ for $x>0$. Then
$\lambda_A\leq\lambda_B$, and $J(\pi;\omega)$ is minimized over all
routing maps for $(F,h)$ by allowing $\bw$ if
$\lambda(\bw)<\lambda_A$, referring it if
$\lambda_A<\lambda(\bw)<\lambda_B$, and blocking it if
$\lambda(\bw)>\lambda_B$. The interval $(\lambda_A,\lambda_B)$ is
nonempty if and only if
\begin{equation}\label{eq:reviewerworth}
c_h<(\kappa_1-\kappa_0)\frac{\ell_M\ell_B}{\ell_M+\ell_B}.
\end{equation}
\end{theorem}
The proof is given in Section~S2.3 of the Supplementary Material. \Cref{thm:routingstructure} answers the three questions posed
above. First, the reviewer is used for patterns whose likelihood
ratio is intermediate. A pattern with $\lambda(\bw)>\lambda_B$ is
strong evidence of an unauthorized action and is blocked
automatically, and a pattern with $\lambda(\bw)<\lambda_A$ is
strong evidence of an authorized action and is allowed
automatically; only the ambiguous patterns between the two
thresholds justify the reviewer's cost.  Second, by
Eq.~\eqref{eq:reviewerworth}, the reviewer can be worth invoking for
some pattern only if its cost is below
$(\kappa_1-\kappa_0)\ell_M\ell_B/(\ell_M+\ell_B)$.  Third, dependence among the first-stage
safeguards leaves the thresholds unchanged and enters only through
the likelihood ratios.  

The following example illustrates \Cref{thm:routingstructure}.

\begin{example}\rm
\label{ex:routingdependence}
Consider two first-stage safeguards and a reviewer, with
$\omega=1/2$, $\ell_M=\ell_B=1$, and $c_h=0.2$. Assume $\kappa_1=0.9$
and $\kappa_0=0.1$. The pattern losses in
Eq.~\eqref{eq:patterncost}, divided by $s_0(\bw)$, are
\[
\frac{\varphi_{\cA}(\bw)}{s_0(\bw)}=\frac{\lambda(\bw)}{2},\qquad
\frac{\varphi_{\cQ}(\bw)}{s_0(\bw)}=0.15\{1+\lambda(\bw)\},\qquad
\frac{\varphi_{\cB}(\bw)}{s_0(\bw)}=\frac12,
\]
and Eq.~\eqref{eq:thresholds} gives $\lambda_A=3/7$ and $\lambda_B=7/3$.
Condition Eq.~\eqref{eq:reviewerworth} holds because $c_h=0.2<0.4$.
These quantities depend on $\omega$, the losses, $c_h$, and
$\kappa_y$, but not on the joint distribution of the first-stage
decisions.

Under Model~1, the first-stage patterns have class-conditional
probabilities
\[
\begin{array}{c|cccc}
\bw & 00 & 01 & 10 & 11\\ \hline
s_0(\bw) & 0.40 & 0.20 & 0.20 & 0.20\\
s_1(\bw) & 0.08 & 0.12 & 0.20 & 0.60
\end{array}
\]
with likelihood ratios $0.2$, $0.6$, $1$, and $3$. The map
minimizing $J(\pi;\omega)$ therefore allows $00$, refers $01$ and
$10$, and blocks $11$. Model~2 is obtained by adding $0.09$ to the probabilities of $00$
and $11$ and subtracting $0.09$ from those of $01$ and $10$,
separately in each authorization class, with the reviewer
unchanged:
\[
\begin{array}{c|cccc}
\bw & 00 & 01 & 10 & 11\\ \hline
s_0(\bw) & 0.49 & 0.11 & 0.11 & 0.29\\
s_1(\bw) & 0.17 & 0.03 & 0.11 & 0.69
\end{array}
\]
This transfer preserves every safeguard's BLOCK probability and
increases the distribution of $(W,H)$ in the supermodular order of
Section~\ref{subsec:stochreview}. Because it acts only on $W$,
$\kappa_0$ and $\kappa_1$, and hence the thresholds, are unchanged.
The likelihood ratio of $01$ falls from $0.6$ to
$0.03/0.11=3/11<\lambda_A$, so its optimal assignment changes from
REVIEW to automatic ALLOW. Patterns $00$ and $10$ keep their
assignments, and $11$ remains blocked because
$0.69/0.29\approx2.38>\lambda_B$; see
Figure~\ref{fig:routingdependencethresholds}.
\begin{figure}[htb!]
\centering
\includegraphics[width=.75\linewidth]{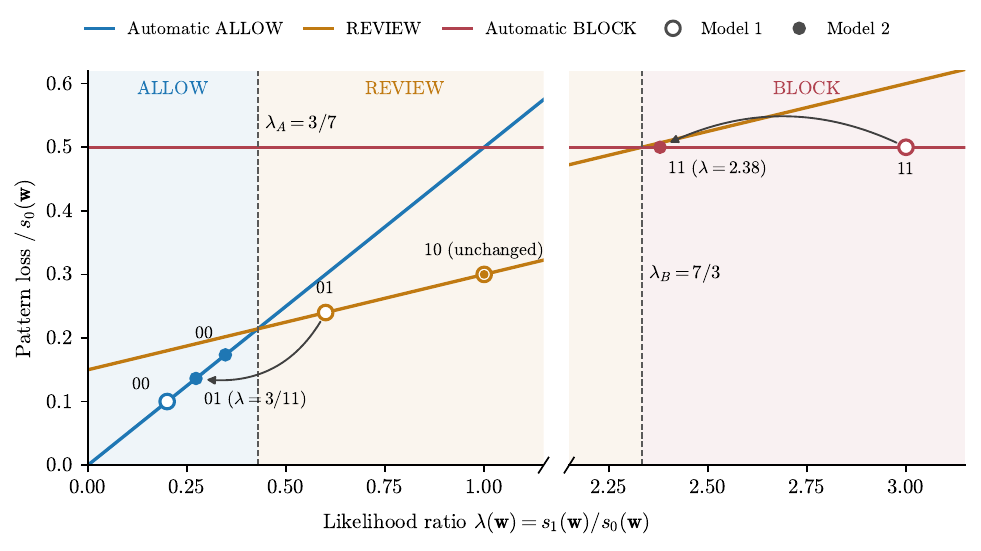}
\caption{Pattern losses, divided by $s_0(\bw)$, and likelihood-ratio
thresholds. Hollow and filled markers
show each pattern under Models~1 and~2, colored by its optimal
assignment. The transfer leaves the component margins and the
thresholds unchanged, but moves pattern $01$ from the REVIEW region
to the automatic-ALLOW region. The horizontal axis is broken between
$1.15$ and $2.15$.}
\label{fig:routingdependencethresholds}
\end{figure}

\end{example}

Algorithm~\ref{alg:jointdesign} searches an optimal policy over both policy classes, i.e.,  sequential layered policies and selective verification policies. Note that Algorithm~\ref{alg:layeredexact} returns the optimum over
admissible sequential layered policies. For each admissible
first-stage reviewer pair $(F,h)$, the binary program optimizes
over all routing rules for that pair. Comparing these optima
therefore gives a globally optimal policy within the specified
candidate classes for the supplied $q_0$ and $q_1$, provided
every routing program is solved to global optimality or
established infeasible.

\begin{algorithm}[htb!]
\caption{Optimal safeguard policy and its risk and cost metrics}
\label{alg:jointdesign}
\small\setstretch{1.0}
\begin{algorithmic}[1]
\Require $q_0,q_1$, $(c_j)_{j=1}^m$, $\omega$, $\beta$,
$O\in\{J,M\}$, $\mathcal S_{\rm adm}$, $\mathcal H_{\rm adm}$;
$\alpha,\ell_M,\ell_B$ if $O=J$, and $b$ if $O=M$
\Ensure An optimal policy and its metrics, \textsc{Infeasible},
or \textsc{Unresolved}
 
\State Compute $a_y(S)$ from Eq.~\eqref{eq:theoryincrement}
\State Run Algorithm~\ref{alg:layeredexact}; let $r_L$ be its output
\State $\mathrm{found}\gets\mathrm{false}$
\If{$r_L=(\pi_L^*,v_L^*)$}
  \State $(\pi^*,v^*)\gets(\pi_L^*,v_L^*)$
  \State $\mathrm{found}\gets\mathrm{true}$
\EndIf
 
\ForAll{$(F,h)\in\mathcal H_{\rm adm}$ in a fixed order}
  \State Compute $s_y(\mathbf w)$ and $t_y(\mathbf w)$ from
         Eq.~\eqref{eq:theoryroutingprob}
  \If{$O=J$}
    \State $r\gets\operatorname{Solve}[\text{Eq.}~Eq.~\eqref{eq:routingMILP}]$
  \Else
    \State $r\gets\operatorname{Solve}[\text{Eq.}~Eq.~\eqref{eq:routingMILPmiss}]$
  \EndIf
  \If{$r=\textsc{Unresolved}$}
    \State \Return \textsc{Unresolved}
  \EndIf
  \If{$r=(\mathcal A,\mathcal B,\mathcal Q)$}
    \State $\pi_{F,h}\gets(F,h,\mathcal A,\mathcal B,\mathcal Q)$
    \State $v_{F,h}\gets J(\pi_{F,h};\omega)$ if $O=J$;
           otherwise $v_{F,h}\gets M(\pi_{F,h})$
    \If{$\mathrm{found}=\mathrm{false}$}
      \State $(\pi^*,v^*)\gets(\pi_{F,h},v_{F,h})$
      \State $\mathrm{found}\gets\mathrm{true}$
    \Else
      \If{$v_{F,h}<v^*$}
        \State $(\pi^*,v^*)\gets(\pi_{F,h},v_{F,h})$
      \EndIf
    \EndIf
  \EndIf
\EndFor
 
\If{$\mathrm{found}=\mathrm{false}$}
  \State \Return \textsc{Infeasible}
\EndIf
\State \Return
$(\pi^*,M(\pi^*),B(\pi^*),K(\pi^*;\omega),v^*)$
\end{algorithmic}
\end{algorithm}
Given $q_0$ and $q_1$, the layered dynamic program
takes $O(m2^m)$ time. The selective verification search considers
at most $m(2^{m-1}-1)$ nonempty first-stage--reviewer pairs
$(F,h)$ with $h\notin F$. For a pair with $f=|F|$, the routing
program has $2^f$ first-stage patterns and $2^{f+1}$ binary
variables. Since each pattern has three routing choices, there
are $3^{2^f}$ possible routing rules. For example, if $F=\{1,2\}$ and $h=3$, then $f=2$.
The four first-stage patterns yield eight binary variables
and $3^4=81$ possible routing rules. These procedures support practical use with safeguard pools
of modest size.The policy search is performed once, offline, on calibration data;
at deployment, each action is screened by the selected policy
alone, so the cost of the search does not affect per-action
screening time.

\section{Numerical Study}
\label{sec:numerical}

This section examines the effects of policy flexibility and
dependence information on safeguard design. To assess policy flexibility, under each specified
dependence model, we compare the optimal feasible policy in the baseline
class (i.e., sequential layered screening and flag-to-review cascades) with
the optimal feasible policy in the full class.

\begin{table}[htb!]
\centering\small
\setlength{\tabcolsep}{4pt}
\caption{Specified heterogeneous safeguard pool. }
\label{tab:components}
\begin{tabular}{crrr}
\toprule
Safeguard & Miss probability $1-p_{j1}$ & False block $p_{j0}$ & $c_j$\\
\midrule
$D_1$ & 0.160 & 0.030 & 0.20 \\
$D_2$ & 0.080 & 0.120 & 0.35 \\
$D_3$ & 0.100 & 0.050 & 0.65 \\
$D_4$ & 0.035 & 0.025 & 2.00 \\
\bottomrule
\end{tabular}
\end{table}
We use four safeguards with the distinct marginal error probabilities
and invocation costs in Table~\ref{tab:components}. Safeguard $D_1$
is the least costly but has the largest miss probability. Safeguard
$D_2$ detects more unauthorized actions, at the expense of more false
blocks; $D_3$ provides a different security--service trade-off.
Safeguard $D_4$ has the smallest error probabilities and the highest
invocation cost.   No component is designated
as the reviewer in advance: every safeguard may be selected as a
screen or as a reviewer in an admissible configuration.
We examine two optimization problems with $\omega=0.10$. The
minimum-total-cost problem Eq.~\eqref{eq:designcost} imposes
$(\alpha,\beta)=(0.05,0.10)$ and uses
$(\ell_M,\ell_B)=(100,10)$, giving
\begin{equation}\label{eq:J}
J(\pi;0.10)=K(\pi;0.10)+10M(\pi)+9B(\pi).
\end{equation}
At operating budget $b=2$, the minimum-miss problem
Eq.~\eqref{eq:designmiss} becomes
\begin{equation}\label{eq:numericalmiss}
\begin{aligned}
\min_{\pi}\quad & M(\pi)\\
\text{subject to}\quad
& B(\pi)\leq 0.10,\quad K(\pi;0.10)\leq 2.
\end{aligned}
\end{equation}

Table~\ref{tab:design} specifies eight class-conditional dependence
scenarios.  S0 is the  independence model. S1 uses a Gaussian copula with
moderate dependence in both classes measured by Kendall's $\tau_y$. S2 and S3 reverse which class has
stronger dependence, using
$(\tau_0,\tau_1)=(0.10,0.50)$ and $(0.50,0.10)$, respectively.
S4--S6 replace the Gaussian copula in S1 with Student-$t$, Clayton,
and Gumbel copulas while holding $\tau_0=\tau_1=0.30$. For the Gaussian and Student-$t$ scenarios S1--S4, we set homogeneous correlation
$\rho_y=\sin(\pi\tau_y/2)$; the Student-$t$ degrees of freedom are fixed
at five. The Clayton and Gumbel parameters are
$\theta_y=2\tau_y/(1-\tau_y)$ and
$\theta_y=1/(1-\tau_y)$, respectively.  
\begin{table}[!htbp]
\centering\small
\setlength{\tabcolsep}{4pt}
\caption{Dependence specifications with the same margins and costs.}
\label{tab:design}
\begin{tabular}{clcc@{\hspace{2.5em}}clcc}
\toprule
Scenario & Copula & $\tau_0$ & $\tau_1$
 & Scenario & Copula & $\tau_0$ & $\tau_1$\\
\midrule
S0 & Independence  & 0.00 & 0.00 & S4 & Student-$t_5$           & 0.30 & 0.30\\
S1 & Gaussian      & 0.30 & 0.30 & S5 & Clayton                 & 0.30 & 0.30\\
S2 & Gaussian      & 0.10 & 0.50 & S6 & Gumbel                  & 0.30 & 0.30\\
S3 & Gaussian      & 0.50 & 0.10 & S7 & Gaussian, pair-specific & --   & --  \\
\bottomrule
\end{tabular}
\end{table}
 S7 uses
pair-specific Gaussian correlations to examine dependence that varies
across safeguard pairs.   Its Gaussian
correlation matrices are specified by
\begin{equation*}
\begin{aligned}
\Sigma_y&=\operatorname{diag}(1-b_{1y}^2,\ldots,1-b_{4y}^2)
                 +\mathbf b_y\mathbf b_y^\top,\\
\mathbf b_0&=(0.80,0.70,0.30,0.55)^\top,\qquad
\mathbf b_1=(0.35,0.85,0.75,0.45)^\top.
\end{aligned}
\end{equation*}

The baseline optimizes over every nonempty ordered layered-screening
policy and every flag-to-review cascade with nonempty first-stage set
$F$ and reviewer $h\notin F$. The full class retains the layered
policies and permits each first-stage pattern to be automatically
allowed, blocked, or referred to the reviewer. As specified in
Section~\ref{subsec:policies}, all safeguards in $F$ are invoked
before routing. For this four-safeguard pool, there are 64 ordered layered policies.
For each of the four possible reviewers, any nonempty subset of the
other three safeguards can serve as $F$, giving
$4(2^3-1)=28$ eligible $(F,h)$ pairs. Enumerating the
ALLOW/BLOCK/REVIEW routing maps for these pairs gives 27,324
selective verification specifications and hence 27,388 candidate
configurations in the full class. 

\subsection{Minimizing total cost }

For each scenario S0--S7, the search uses the distribution in Table~\ref{tab:design},
with safeguard margins and costs held fixed. For the objective in  Eq.~\eqref{eq:J}, the full search selects a
selective verification policy in each of the eight scenarios, although
sequential layered policies are also eligible. 
\begin{table}[htb!]
\centering\small
\setlength{\tabcolsep}{4pt}
\caption{Distinct verification policies selected by the primary
minimum-total-cost search. Patterns list decisions in increasing
safeguard order within $F$; 0 denotes ALLOW and 1 denotes BLOCK.
Patterns outside $\mathcal B$ and $\mathcal Q$ are automatically
allowed. All safeguards in $F$ are invoked before routing.}
\label{tab:returnedroutes}
\begin{tabular}{llccll}
\toprule
Policy & Selected in & $F$ & $h$ & $\mathcal B$ & $\mathcal Q$\\
\midrule
R1 & S0       & $\{1,2\}$   & $3$ & $\{11\}$     & $\{10,01\}$ \\
R2 & S1,S3,S6 & $\{1,2\}$   & $4$ & $\{11\}$     & $\{10,01\}$ \\
R3 & S4       & $\{1,2\}$   & $4$ & $\{10,11\}$  & $\{01\}$ \\
R4 & S2,S5    & $\{1,2,3\}$ & $4$ & $\{101,111\}$&
     $\{100,010,110,001,011\}$ \\
R5 & S7       & $\{1,2\}$   & $3$ & $\{10,11\}$  & $\{01\}$ \\
\bottomrule
\end{tabular}
\end{table}
Table~\ref{tab:returnedroutes} lists the five distinct
selective verification policies selected across the eight scenarios.
They use three routing maps. R1 and R2 automatically block pattern
$11$ and refer the single-flag patterns $10$ and $01$, but use
different reviewers. R3 and R5 instead automatically block $10$
and $11$ and refer only $01$, again differing in reviewer. R4
selects a larger first-stage set and refers five of its eight
decision patterns. All five policies automatically allow the
all-ALLOW pattern. Thus, changing the joint decision distribution
can change not only which reviewer is selected but also which
first-stage patterns are blocked or referred, even though the
component margins and invocation costs remain fixed across scenarios.
The final-decision rule of each of R1--R5 is neither
supermodular nor submodular; for R1 and R2, for instance,
$\Delta_{12}G_\pi=1$ when the reviewer allows, whereas
$\Delta_{1H}G_\pi=-1$ when $D_2=1$. By \Cref{thm:selectivesm}(a),
the direction of their response to a change in dependence is
therefore not determined in advance; the independence comparison
below shows that it is upward for misses in every dependent
scenario.

To assess the value of the proposed policy class, we compare its
optimized designs with  sequential
layered screening and flag-to-review cascading. For each scenario,
the baseline is the best feasible policy across those two
architectures, while the full search additionally allows
pattern-specific selective verification. Both searches use the same
joint distribution, costs, and risk constraints. Table~
\ref{tab:optimizeddesign}   measures the improvement
attributable to flexible routing. It    compares the selected policy with
the best feasible baseline in each scenario.

\begin{table}[htb!]
\centering\scriptsize
\setlength{\tabcolsep}{2.5pt}
\caption{Best feasible baseline and full-class policy in each scenario
under $M\leq0.05$, $B\leq0.10$, and
$(\omega,\ell_M,\ell_B)=(0.10,100,10)$. Both are optimized using
the scenario's specified joint distribution. The baseline class
contains sequential layered screening and flag-to-review cascading;
the full class additionally permits selective verification.
Policies R1--R5 are defined in Table~\ref{tab:returnedroutes}.}
\label{tab:optimizeddesign}
\begin{tabular}{lcrrrrcrrrr}
\toprule
& \multicolumn{5}{c}{Best feasible baseline}
& \multicolumn{5}{c}{Full-class selection}\\
\cmidrule(lr){2-6}\cmidrule(l){7-11}
Scenario & Policy & $M$ & $B$ & $K$ & $J$
         & Policy & $M$ & $B$ & $K$ & $J$\\
\midrule
S0 & $C(1,2;4)$ & 0.0474 & 0.0037 & 1.0110 & 1.5174
   & R1 & 0.0342 & 0.0107 & 0.6475 & 1.0865\\
S1 & $L(1,3)$ & 0.0413 & 0.0726 & 0.7779 & 1.8441
   & R2 & 0.0488 & 0.0230 & 0.8071 & 1.5017\\
S2 & $L(4)$ & 0.0350 & 0.0250 & 2.0000 & 2.5750
   & R4 & 0.0490 & 0.0105 & 1.5555 & 2.1399\\
S3 & $L(1,3)$ & 0.0233 & 0.0658 & 0.7779 & 1.6032
   & R2 & 0.0297 & 0.0326 & 0.7831 & 1.3740\\
S4 & $L(1,3)$ & 0.0460 & 0.0693 & 0.7779 & 1.8612
   & R3 & 0.0485 & 0.0385 & 0.7617 & 1.5938\\
S5 & $L(4)$ & 0.0350 & 0.0250 & 2.0000 & 2.5750
   & R4 & 0.0490 & 0.0102 & 1.5550 & 2.1369\\
S6 & $L(1,3)$ & 0.0345 & 0.0651 & 0.7779 & 1.7084
   & R2 & 0.0432 & 0.0275 & 0.7843 & 1.4635\\
S7 & $L(1,3)$ & 0.0291 & 0.0761 & 0.7779 & 1.7541
   & R5 & 0.0418 & 0.0388 & 0.6194 & 1.3873\\
\bottomrule
\end{tabular}
\end{table}
  In S0, the best baseline is the flag-to-review cascade
$C(1,2;4)$, while the full search selects R1. The best baseline is
layered screening in every other scenario: $L(1,3)$ in S1, S3,
S4, S6, and S7, and the single safeguard $L(4)$ in S2 and S5.
The full search reduces total expected cost by 14.3\%--28.4\%
relative to these scenario-specific baselines.   It should be noted that the selected
policy need not improve every metric. In S1, for example, R2 has
a higher miss probability than the layered baseline
($0.0488$ versus $0.0413$), but a substantially lower false-block
probability ($0.0230$ versus $0.0726$). Both satisfy the risk
constraints, and the reduction in false-block loss yields a lower
total cost ($1.5017$ versus $1.8441$).

\paragraph{Dependence versus independence models.}

To assess the value of dependence information, we compare the
policy selected under each scenario's dependence model with
the policy selected under independence (R1, selected in S0 in
Table~\ref{tab:optimizeddesign}).   We evaluate this fixed
independence-selected policy and each scenario's dependence-aware
selection under the same scenario-specific joint distribution,
holding the policy class, costs, and risk requirements fixed.

\begin{figure}[htb!]
\centering
\includegraphics[width=.7\linewidth]
  {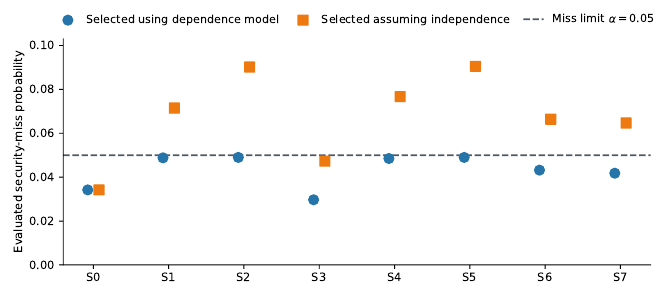}
\caption{Security-miss probabilities of total-cost policies selected
using either the scenario's dependence model (circles) or its
same-margin independence model (squares). 
The dashed line marks $\alpha=0.05$.}
\label{fig:dependencedesign}
\end{figure}

When the selected policies are evaluated under each scenario's
specified dependence model, R1 exceeds the miss limit in S1,
S2, and S4--S7 (Figure~\ref{fig:dependencedesign}). Its miss
probability reaches $0.0905$ in S5, compared with the limit
$\alpha=0.05$. The dependence-aware search selects a feasible
policy in every scenario. In S1, S3, and S6, it selects R2
instead of R1: the first-stage safeguards and routing rule remain
the same, but the reviewer changes from $D_3$ to $D_4$.
Thus, accounting for dependence can change reviewer selection
without changing the first-stage architecture.

S3 distinguishes loss of optimality from loss of feasibility.
The independence selected R1 remains feasible but has
$J=1.4918$, compared with $J=1.3740$ for the dependence-aware
R2, a reduction of $7.9\%$. R2's higher operating cost
($0.7831$ versus $0.6258$) is offset by lower error losses.
In S2 and S5, the independence-selected policy has a
 smaller $J$ than the dependence-aware optimum,
but violates the miss constraint. It is therefore not a
cheaper feasible alternative.

\subsection{Security under operating budgets}

We now consider the minimum-miss problem
Eq.~\eqref{eq:numericalmiss}.   Since the cost of $D_4$ is $c_4=2$ in Table \ref{tab:components}, this budget
permits a policy that invokes $D_4$ for every candidate.
Table~\ref{tab:budgetdesign} compares the best feasible
baseline and full-class policies. It also reports the miss
probability of the policy selected under independence, with
all policies evaluated under each scenario's dependence model.

\begin{table}[htb!]
\centering\small
\setlength{\tabcolsep}{4pt}
\caption{Minimum-miss designs under $K(\pi;0.10)\leq2$ and
$B(\pi)\leq0.10$.  $M_{\rm ind}$ is the miss
probability of the independence-selected policy R6. All
reported metrics are evaluated under the scenario's dependence model.}
\label{tab:budgetdesign}
\begin{tabular}{lcrrcrrr}
\toprule
& \multicolumn{3}{c}{Best feasible baseline}
& \multicolumn{3}{c}{Full-class selection} & \\
\cmidrule(lr){2-4}\cmidrule(lr){5-7}
Scenario & Policy & $M$ & $K$ & Policy & $M$ & $K$
         & $M_{\rm ind}$\\
\midrule
S0 & $L(1,4)$ & 0.0056 & 1.9780
   & R6 & 0.0018 & 1.4020 & 0.0018\\
S1 & $L(1,4)$ & 0.0176 & 1.9780
   & $L(1,4)$ & 0.0176 & 1.9780 & 0.0189\\
S2 & $L(1,4)$ & 0.0271 & 1.9780
   & $L(1,4)$ & 0.0271 & 1.9780 & 0.0382\\
S3 & $L(1,4)$ & 0.0090 & 1.9780
   & R6 & 0.0054 & 1.3451 & 0.0054\\
S4 & $L(1,4)$ & 0.0201 & 1.9780
   & $L(1,4)$ & 0.0201 & 1.9780 & 0.0235\\
S5 & $L(1,4)$ & 0.0279 & 1.9780
   & $L(1,4)$ & 0.0279 & 1.9780 & 0.0380\\
S6 & $L(1,4)$ & 0.0137 & 1.9780
   & R6 & 0.0112 & 1.3569 & 0.0112\\
S7 & $L(1,4)$ & 0.0090 & 1.9780
   & $L(1,4)$ & 0.0090 & 1.9780 & 0.0144\\
\bottomrule
\end{tabular}
\end{table}

 We compare the best feasible
layered-screening or flag-to-review baseline with the full
policy class, with both optimized under each scenario's joint
distribution. The best baseline is $L(1,4)$ in every scenario. Early stopping
allows this two-component policy to satisfy the budget, with
$K=1.9780$. The full search selects the same policy in five
scenarios but improves the miss probability in S0, S3, and S6
by selecting R6. This policy uses $F=\{1,2,3\}$ and reviewer
$D_4$: it automatically allows $000$, refers $010$, and
automatically blocks every other first-stage pattern. Thus,
the additional routing choices improve security in some
scenarios, but not uniformly across the specified distributions.

The independence-based
search selects R6. Table~\ref{tab:budgetdesign}
evaluates this selection under each scenario's joint
distribution. R6 satisfies the operating-budget and false-block
constraints throughout, but is not always minimum-miss. For
example, in S2 its miss probability is $0.0382$, compared with
$0.0271$ for the dependence-aware selection $L(1,4)$. Dependence
information therefore changes the preferred architecture in
this case even though the independence selected policy remains
feasible.

\section{Empirical Study}
\label{sec:empirical}\label{sec:e1recomp}
\begin{figure}[htb!]
\centering
\includegraphics[width=.95\textwidth]{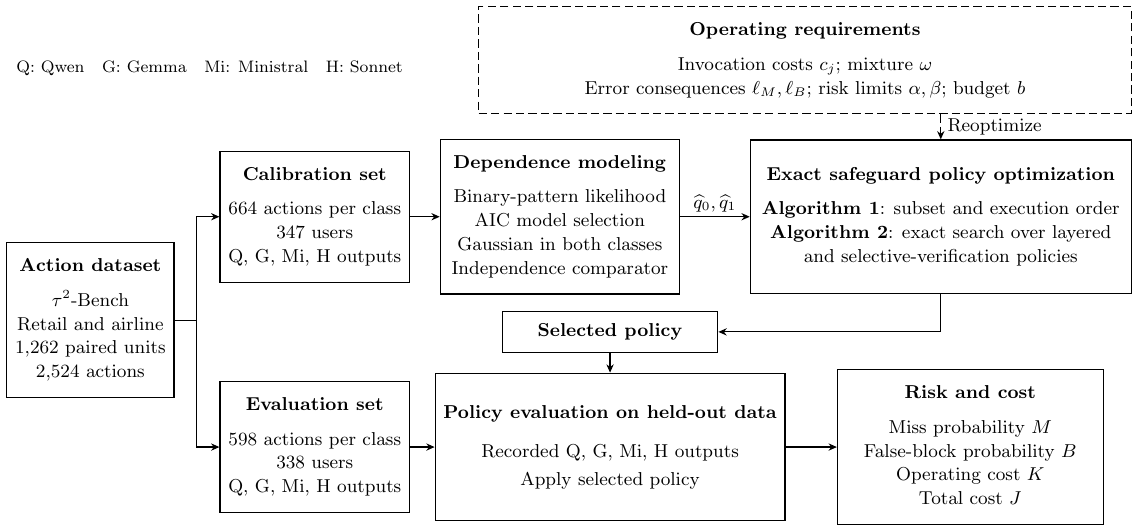}
\caption{Overview of dependence-aware
safeguard risk assessment and optimal policy selection.}
\label{fig:empirical-workflow}
\end{figure}

In this section, we evaluate the proposed framework using an empirical
dataset of authorized and unauthorized agent actions. We investigate
how dependence among safeguard decisions affects system risk
and policy selection, and assess the effectiveness of DAPOA in identifying safeguard configurations under
specified risk and cost requirements.
Figure~\ref{fig:empirical-workflow} illustrates the overall empirical
procedure, including dependence modeling, optimal policy selection,
and evaluation on held-out  data.

\subsection{Dataset and experimental implementation}
\label{subsec:empiricaldata}

The studied data are constructed from the retail and airline environments of
$\tau^2$-Bench \citep{barres2026tau2}, a benchmark for conversational
agents that interact with simulated users and database-backed tools
under domain-specific policies. These environments represent
customer-service operations in which a tool call must satisfy both
business rules and the user's request. Retail cases are generated from
six request templates and sandbox tool responses, while airline cases
cover reservations and bookings, baggage, flight changes, passenger names,
cancellations, and certificates. Each unit contains a trusted interaction
context together with one authorized action and one unauthorized variant.
The authorized action is the confirmed sandbox call, whereas the
unauthorized action is obtained by the first applicable perturbation of
eligibility, cabin, scope, target, refund destination, recipient, or reason.
Labels follow from this construction and the trusted authorization
specification, independently of safeguard outputs.

The dataset comprises 1,262 paired units: 555 retail orders and 707
airline cases, giving 2,524 candidate actions. We divide the data into
a calibration set and an evaluation set so that dependence modeling
and safeguard-policy selection are carried out without using the
actions on which the resulting policies are subsequently assessed.
The calibration set is used to estimate the
marginal error rates and dependence structure and to select the
safeguard configuration, whereas the evaluation set is used to
assess the selected policy. The split is made at the user level and
stratified by source environment; all orders or reservations belonging
to the same user, together with both actions in each pair, remain in
the same set. Calibration contains 664 authorized and 664 unauthorized
actions from 347 users, and evaluation contains 598 actions of each
class from 338 different users; see Table \ref{tab:emp-dataset}. This grouping prevents user-specific
information from appearing in both model development and evaluation.
\begin{table}[htb!]
\centering\small
\caption{Composition of the action-authorization dataset. Each paired
unit contains one authorized and one unauthorized action. Users are
disjoint between calibration and evaluation.}
\label{tab:emp-dataset}
\begin{tabular}{lrrrr}
\toprule
Set&Users&Paired units&Authorized&Unauthorized\\
\midrule
Calibration&347&664&664&664\\
Evaluation&338&598&598&598\\
\midrule
Total&685&1,262&1,262&1,262\\
\bottomrule
\end{tabular}
\end{table}

Each safeguard is evaluated independently on the same candidate action
and trusted context. The prompt asks whether the proposed action is
authorized under the specification and requires a
response with an \texttt{ALLOW} or \texttt{BLOCK} decision and a short
explanation.

The empirical study uses four LLM-based authorization safeguards:
Qwen3 8B (Q), Gemma4 E4B (G), Ministral 3 8B (Mi), and Claude
Sonnet 5 (H). The three local models are evaluated through Ollama using
the identifiers \texttt{qwen3:8b}, \texttt{gemma4:e4b}, and
\texttt{ministral-3:8b}. Local inference is performed on a Mac with an
Apple M4 Pro processor and 24 GB of unified memory. Sonnet is accessed
through the Claude Code command-line interface. The local models are
evaluated at temperature zero with an 8,192-token context allocation;
the maximum recorded local input length is 4,724 tokens.
Additional implementation details are provided in
Section~S3 of the Supplementary Material.

To compare policies with different invocation patterns, we assign each
safeguard a fixed normalized invocation cost. Local inference is valued
at an assumed rate of \$0.20 per million input-plus-output tokens.
Although Sonnet responses were obtained through a subscription
interface, they are valued using the API-equivalent rates, namely
\$2 per million input tokens and \$10 per million output tokens
\citep{anthropic2026pricing}. These rates provide a common
accounting basis; they are not the actual expenditure incurred in
collecting the responses. Qwen is normalized to
$c_Q=0.10$, and the costs of the remaining safeguards are determined
from calibration-based call usage under the specified token-price
convention. We retain the fixed baseline coefficients reported in
Table~\ref{tab:emp-components} throughout the dataset.
The consequence losses $\ell_M$ and $\ell_B$ use the same normalized
cost unit and are illustrative decision inputs.
Details of the cost construction are provided in
Section~S3 of the Supplementary Material.

\begin{table}[htb!]
\centering\small
\caption{Component errors in the calibration sample and
fixed baseline invocation costs. Error counts are based
on 664 actions per authorization class.}
\label{tab:emp-components}
\begin{tabular}{lrrr}
\toprule
Component & $c_j$ & Misses & False blocks\\
\midrule
Qwen (Q)      & 0.1000 & 124 & 9\\
Gemma (G)     & 0.1040 & 172 & 2\\
Ministral (Mi)& 0.1020 & 13  & 20\\
Sonnet (H)    & 1.2840 & 2   & 0\\
\bottomrule
\end{tabular}
\end{table}
Table~\ref{tab:emp-components} reports the marginal error
counts of the four safeguards in the calibration sample,
together with their normalized invocation costs.
The safeguards differ substantially in their marginal
performance. Sonnet has the fewest calibration misses and no observed
calibration false blocks, but incurs the highest invocation cost.
These calibration outcomes are used to estimate the
marginal distributions and dependence
structure for subsequent policy selection.

For unauthorized actions, the blocking proportions for
$(Q,G,\mathrm{Mi},H)$ are
\[
(540/664,\;492/664,\;651/664,\;662/664).
\]
The corresponding proportions for authorized actions are
$(9/664,2/664,20/664,0).$
Dependence estimation therefore uses all four safeguards for
unauthorized actions and the three varying local safeguards for
authorized actions. Sonnet's authorized-class decision is constant
in calibration and is attached through its empirical margin.
\begin{table}[htb!]
\centering\small
\caption{Sample Kendall's $\tau$ between local safeguards'
binary decisions in the calibration sample. Each authorization
class contains 664 actions.}
\label{tab:emp-dependence}
\begin{tabular}{lrr}
\toprule
Pair & Unauthorized & Authorized\\
\midrule
$(Q,G)$           & 0.431 & $-0.006$\\
$(Q,\mathrm{Mi})$ & 0.211 & 0.208\\
$(G,\mathrm{Mi})$ & 0.214 & 0.151\\
\bottomrule
\end{tabular}
\end{table}
Table~\ref{tab:emp-dependence} reports the sample pairwise
associations separately for the two authorization classes.
For unauthorized actions, all three coefficients are positive,
with the strongest association between Qwen and Gemma.
For authorized actions, the Qwen--Gemma association is close
to zero, whereas both pairs involving Ministral remain
positively associated. The Qwen--Ministral association is
similar across classes, while the other two pairs exhibit
different association strengths. For unauthorized actions,
positive association also characterizes the corresponding
security-miss indicators; for authorized actions, BLOCK
decisions are themselves false-block indicators. These
observations reveal different patterns of error overlap
across safeguard pairs and authorization classes.

To accommodate different pairwise strengths, we fit Gaussian and
Student-$t$ copulas. The exchangeable multivariate Clayton and Gumbel
specifications  do not allow these
distinct pairwise strengths. With margins fixed at their calibration
proportions, dependence parameters are estimated by maximizing the
likelihood of the observed binary decision patterns.
Model selection is based on the Akaike information criterion (AIC),
including independence as a candidate. AIC selects the Gaussian
copula for both classes: the Gaussian, Student-$t$, and independence
values are $1412.28$, $1414.80$, and $1554.30$ for unauthorized actions,
and $292.98$, $295.01$, and $302.00$ for authorized actions.
Further details of the estimation and model-selection procedures
are provided in Section~S3 of the Supplementary Material.

We consider two classes of safeguard policies. The baseline class
includes sequential layered screening with different safeguard subsets
and invocation orders, together with flag-to-review cascades. The full
class additionally allows flexible ALLOW/BLOCK/REVIEW routing based on
the first-stage decisions. With four safeguards, the full class contains
64 ordered layered-screening policies and 27,324 selective verification
policies across 28 first-stage reviewer combinations, giving a total
of 27,388 candidate configurations. All four single-safeguard policies
are included. For the principal analysis, we set $\omega=0.50$,
$(\ell_M,\ell_B)=(100,10)$, and
$(\alpha,\beta)=(0.05,0.10)$, with invocation costs given in
Table~\ref{tab:emp-components}. We also consider the minimum-miss
problem with $\beta=0.10$ and operating budgets
$b\in\{0.25,0.50,1.00\}$. For each operating setting,
DAPOA is applied to select the optimal safeguard policy under the
fitted copula model and the independence assumption.
Comparing policies selected under the same
requirements allows us to assess the effects of dependence modeling
and the additional flexibility of selective verification.
The authorization labels are used for calibration and evaluation;
the selected policies operate using only observed safeguard decisions.
Additional implementation details are provided in
Section~S3 of the Supplementary Material.

\subsection{Impact of dependence on safeguard design}
\label{subsec:empiricaldependence}
 
Table~\ref{tab:emp-keypolicies} reports the policies selected under
independence and under Gaussian dependence modeling, together with
their evaluation performance, for the principal setting and four
variations. In the table, $\pi_{\mathrm I}$ invokes $Q$, $G$, and
$\mathrm{Mi}$, allows when all three allow, blocks when at least two
block, and otherwise refers to Sonnet. The policy $\pi_{\mathrm V}$
invokes $Q$ and $G$, blocks when both block, and otherwise refers to
Sonnet. In each case Sonnet's decision is final when review is
invoked.
 
\begin{table}[htb!]
\centering\small
\caption{Optimal policy selections in the full class and evaluation
performance under different operating settings. Unless otherwise
specified, parameters take their principal values. Error counts are
based on 598 actions per class. Sequential arrows denote stopping at
the first BLOCK. The dagger indicates a violation of at least one
evaluation risk limit.}
\label{tab:emp-keypolicies}
\setlength{\tabcolsep}{4pt}
\begin{tabular}{lllrrrr}
\toprule
Setting & Model & Selected policy & Misses & False blocks
& $\widehat K$ & $\widehat J$\\
\midrule
$\ell_M=10$
 & Independence & $\mathrm{Mi}\to G$ & 15 & 32 & 0.1529 & 0.5458\\
 & Gaussian & $\mathrm{Mi}$ & 17 & 30 & 0.1020 & 0.4950\\
\addlinespace[3pt]
 
Principal ($\ell_M=100$)
 & Independence & $\pi_{\mathrm I}$ & 14 & 9 & 0.3779 & 1.6237\\
 & Gaussian & $\mathrm{Mi}\to H$ & 0 & 30 & 0.7300 & 0.9809\\
\addlinespace[3pt]
 
$\ell_M=1000$
 & Independence & $\mathrm{Mi}\to Q\to G$ & 14 & 32 & 0.2016 & 12.1749\\
 & Gaussian & $\mathrm{Mi}\to H$ & 0 & 30 & 0.7300 & 0.9809\\
\addlinespace[3pt]
 
$\alpha=0.005,\ \beta=0.01$
 & Independence & $\pi_{\mathrm I}^{\dagger}$ & 14 & 9 & 0.3779 & 1.6237\\
 & Gaussian & $G\to H$ & 0 & 5 & 0.8705 & 0.9123\\
\addlinespace[3pt]
 
$\ell_B=100$
 & Independence & $\pi_{\mathrm I}$ & 14 & 9 & 0.3779 & 2.3010\\
 & Gaussian & $\pi_{\mathrm V}$ & 0 & 3 & 1.0070 & 1.2579\\
\bottomrule
\end{tabular}
\end{table}
 
In the principal setting, the Gaussian-based selection reduces the
number of misses from 14 to zero and total cost by $39.6\%$, while
increasing false blocks from nine to 30. Its operating cost is
higher than that of the independence-selected policy. Relative to
Sonnet alone, which has zero misses, two false blocks, and
$(\widehat K,\widehat J)=(1.2840,1.3007)$, the selected cascade
reduces total cost by $24.6\%$ and invokes Sonnet on 585 of 1,196
evaluation actions rather than on every action.
 
Figure~S1 of the Supplementary Material compares the effects of
security-miss and false-block consequences using the same loss
decomposition. When $\ell_M=10$, the Gaussian model selects
Ministral alone, whereas independence adds Gemma. The former has
two additional observed misses, two fewer false blocks, and lower
operating and total costs. As $\ell_M$ increases to $100$ and
$1000$, the Gaussian-selected policy invokes Sonnet after a
Ministral ALLOW, recording zero evaluation misses. At
$\ell_M=1000$, the independence-based selection has 14 misses and
total cost $12.1749$, compared with $0.9809$ under Gaussian-based
selection. When false blocks instead receive a high consequence,
$\ell_B=100$, the Gaussian model selects $\pi_{\mathrm V}$ and
records three false blocks, compared with 30 at the principal
setting. Its operating cost increases from $0.7300$ to $1.0070$.
These results show how the selected policy balances both error
consequences against safeguard invocation cost. Under the stricter security requirement
$(\alpha,\beta)=(0.005,0.01)$, independence again selects
$\pi_{\mathrm I}$, whose evaluation miss and false-block rates,
$14/598=0.02341$ and $9/598=0.01505$, exceed both specified limits.
In contrast, the Gaussian model selects $G\to H$, which satisfies
both requirements on the evaluation sample, albeit at a higher
operating cost.
 
\begin{figure}[htb!]
\centering
\includegraphics[width=.95\linewidth]{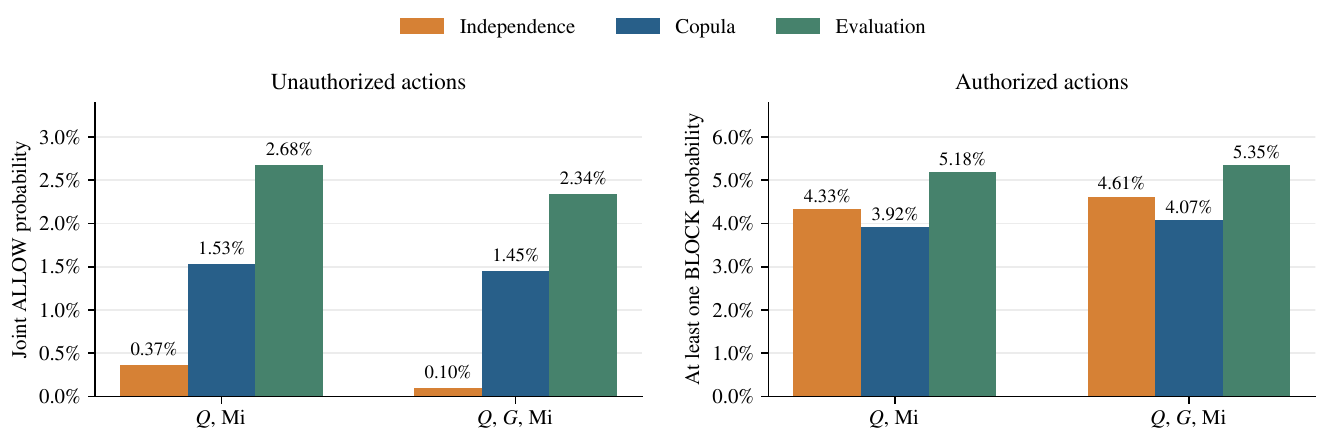}
\caption{Fitted and evaluation probabilities of the error events of
independence-selected policies. Left: joint ALLOW on unauthorized
actions. Right: at least one BLOCK on authorized actions.}
\label{fig:emp-joint}
\end{figure}

\paragraph{Why independence-based selection fails.}
The 14 misses of $\pi_{\mathrm I}$ all occur when $Q$, $G$, and
$\mathrm{Mi}$ simultaneously return ALLOW, the pattern that
$\pi_{\mathrm I}$ allows automatically. The same event produces the
14 misses of $\mathrm{Mi}\to Q\to G$, the independence selection
when $\ell_M=1000$. Similarly, all 16 evaluation misses of
$\pi_{Q,\mathrm{Mi}}$, which the independence model selects in
eight of the nine cost--mixture settings with $\omega=0.10$
(Section~S3.3 of the Supplementary Material), occur when $Q$ and
$\mathrm{Mi}$ both return ALLOW. The miss probabilities of these
policies are therefore governed by joint-ALLOW probabilities, which
Figure~\ref{fig:emp-joint} compares with the predictions of the two
fitted models.
 
For unauthorized actions, independence predicts
$0.0010$ for $\{Q,G,\mathrm{Mi}\}$ and $0.0037$ for
$\{Q,\mathrm{Mi}\}$, fewer than one expected miss in 598 actions,
and underestimates the observed proportions, $14/598=0.0234$ and
$16/598=0.0268$, by factors of about $25$ and $7$. This explains why
the independence model regards these policies as safe, and why it
selects $\pi_{\mathrm I}$ even under the limit $\alpha=0.005$. The
Gaussian predictions, $0.0145$ and $0.0153$, are much closer,
although both remain below the evaluation proportions. For authorized actions, at least one BLOCK by
$\{Q,G,\mathrm{Mi}\}$ is the false-block event of
$\mathrm{Mi}\to Q\to G$, which occurs for 32 of 598 actions
($0.0535$); the corresponding event for $\{Q,\mathrm{Mi}\}$ occurs
for 31 ($0.0518$). Independence predicts $0.0462$ and $0.0433$, and
the Gaussian model $0.0407$ and $0.0392$. Again both models underpredict
these proportions

 
The selected policies also differ in how predictably they respond
to misspecified dependence. The Gaussian-selected policies
$\mathrm{Mi}\to H$ and $G\to H$ block when the first-stage
safeguard blocks and refer its ALLOW decisions to Sonnet, so their
final-decision rule is $D_j\vee H$ for $j\in\{\mathrm{Mi},G\}$,
which is submodular. By \Cref{thm:selectivesm}(a), at fixed margins
a supermodular increase in dependence raises their miss probability
and lowers their false-block probability, so the direction of their
sensitivity is known in advance. The rules of $\pi_{\mathrm I}$ and
$\pi_{\mathrm V}$ are neither supermodular nor submodular; for
$\pi_{\mathrm V}$, $\Delta_{QG}G_\pi=1$ when Sonnet allows, whereas
$\Delta_{QH}G_\pi=-1$ when Gemma blocks. By the same result, the
direction of their response is not determined.

\subsection{Sensitivity analysis}
\label{sec:sens}
 
We examine whether the benefits of dependence-aware safeguard design
persist under different operating conditions, and how the proposed
algorithms balance security risk, service disruption, and invocation costs.
The four safeguards, their class probability distributions,
and the admissible policy classes are held fixed. For each setting, DAPOA selects the optimal policy under both the
Gaussian and the independence models, using the same fitted
distributions, and both selections are evaluated on the held-out
sample.
We consider four groups of settings:
\begin{itemize}
\setlength{\itemsep}{2pt}
\item \emph{Miss consequences and risk requirements.}
The three miss-consequence settings $\ell_M\in\{10,100,1000\}$ under
$(\alpha,\beta)=(0.05,0.10)$ are reported in the preceding subsection.
We additionally examine the risk limits
$(\alpha,\beta)\in\{(0.01,0.10),(0.005,0.01),(0.02,0.003)\}$
with $\ell_M=100$. Throughout, $\ell_B=10$, $\omega=0.50$,
and invocation costs remain at their baseline values.
\item \emph{Operating budgets.}
For minimum-miss design, we impose $K\le b$ with
$b\in\{0.25,0.50,1.00\}$ and $\beta=0.10$;
there is no separate constraint on $M$. Baseline costs and
$\omega=0.50$ are retained.
\item \emph{Invocation costs and unauthorized-action mixture.}
We multiply local costs by $s_L\in\{0.5,1,2\}$ and Sonnet's cost by
$s_H\in\{0.5,1,2\}$, and set
$\omega\in\{0.01,0.10,0.50\}$. All 27 settings minimize $J$ with
$(\alpha,\beta)=(0.05,0.10)$ and $(\ell_M,\ell_B)=(100,10)$.
The original cost normalization and loss units are not changed.
\item \emph{False-block consequences.}
We set $\ell_B\in\{1,100\}$, holding all other inputs at their
principal values.
\end{itemize}
The principal setting belongs to both the first and third groups,
giving $6+3+27+2-1=37$ distinct settings. Because the miss-consequence
comparison was presented above, we focus here on the alternative
risk limits and the remaining sensitivity groups.
 
\begin{figure}[htb!]
\centering
\includegraphics[width=.98\linewidth]{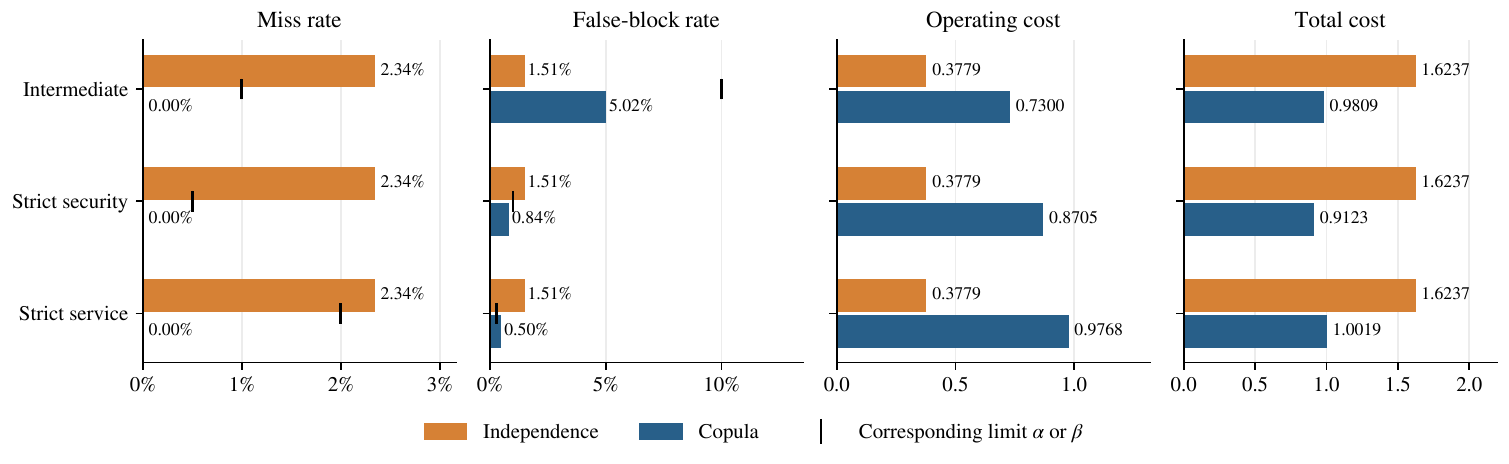}
\caption{Evaluation miss rates, false-block rates, operating costs,
and total costs under three alternative risk requirements. Markers
show the corresponding miss limits $\alpha$ and false-block limits
$\beta$. All settings use baseline costs, $\omega=0.50$, and
$(\ell_M,\ell_B)=(100,10)$.}
\label{fig:emp-reliability-limits}
\end{figure}

\paragraph{Miss consequences and risk requirements.}
Figure~\ref{fig:emp-reliability-limits} compares both evaluation error
rates with their corresponding limits, alongside operating and total
costs, under the three alternative risk requirements. At the intermediate
security limit $(\alpha,\beta)=(0.01,0.10)$, the Independence and
Gaussian models retain their principal selections, $\pi_{\mathrm I}$
and $\mathrm{Mi}\to H$, respectively. The independence-selected
policy has an evaluation miss rate of $14/598=0.02341$, exceeding
$\alpha=0.01$, whereas the Gaussian-selected policy satisfies both
evaluation limits. Under the stricter security limit $(0.005,0.01)$,
the Gaussian selection changes to $G\to H$, recording zero misses
and five false blocks. Its operating cost rises to $0.8705$, while
its total cost is lower than that of the independence-selected policy
($0.9123$ versus $1.6237$).
 
Under the strict service limit $(\alpha,\beta)=(0.02,0.003)$,
the Gaussian model selects flexible verification: Gemma and Ministral
are invoked first, both BLOCK decisions cause an immediate BLOCK,
and every other first-stage pattern is referred to Sonnet.
This policy has zero evaluation misses, three false blocks, and
$(\widehat K,\widehat J)=(0.9768,1.0019)$.
The Independence-based selection $\pi_{\mathrm I}$ has 14 misses,
nine false blocks, and
$(\widehat K,\widehat J)=(0.3779,1.6237)$.
Relative to the Gaussian model's own optimized baseline, Sonnet alone,
flexible verification reduces total cost from $1.3007$ to $1.0019$
and Sonnet invocations from 1,196 to 718, with one additional false
block. However, its evaluation false-block rate $3/598=0.00502$
exceeds $\beta=0.003$; Sonnet alone also exceeds this limit with
$2/598=0.00334$. Thus, neither selected policy satisfies the strict
service requirement on evaluation.
 
\paragraph{Operating budgets.}
For $b=0.25$ and $0.50$, both models select
$\mathrm{Mi}\to Q\to G$, with 14 evaluation misses, 32 false
blocks, and operating cost $0.2016$.
At $b=1.00$, the Gaussian model selects $\mathrm{Mi}\to G\to H$
and the Independence model selects $\mathrm{Mi}\to Q\to G\to H$.
Both policies record zero misses and 32 false blocks, but their
operating costs are $0.7766$ and $0.8243$, respectively.
The Gaussian-based selection avoids a Qwen invocation without
changing the observed final decisions, saving approximately $0.0477$
per action.  
 
\begin{figure}[htb!]
\centering
\includegraphics[width=.9\linewidth]{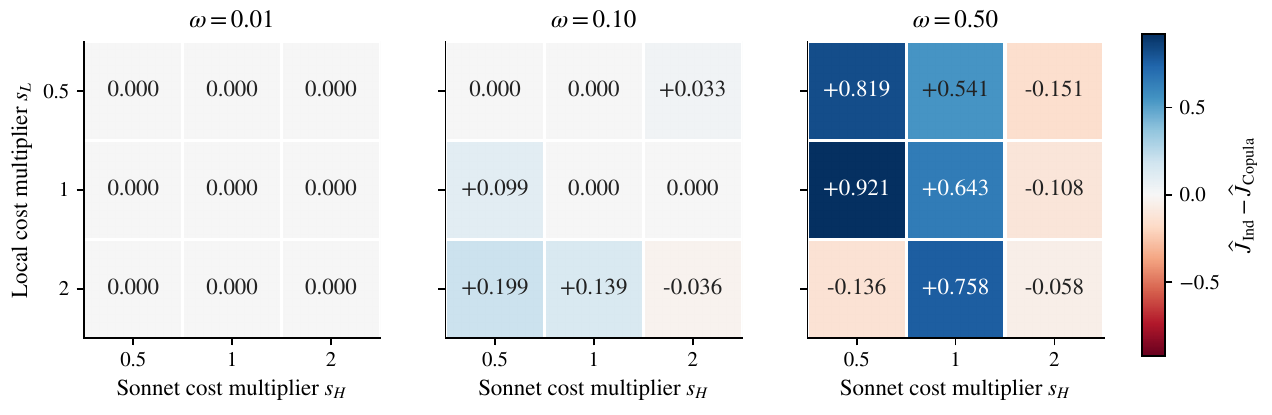}
\caption{Evaluation total-cost difference
$\widehat J_{\mathrm{Independence}}-\widehat J_{\mathrm{Gaussian}}$
across all cost--mixture settings. Positive values favor
Gaussian-based policy selection.}
\label{fig:emp-cost-savings}
\end{figure}

\paragraph{Invocation costs and unauthorized-action mixture.}
Figure~\ref{fig:emp-cost-savings} reports the evaluation total-cost
difference between the Independence and Gaussian selections for the
27 cost--mixture settings (see Section~S3.3 of the Supplementary Material).
At $\omega=0.01$, both models select the same flag-to-review cascade
in all nine cells: Ministral ALLOW decisions are accepted immediately,
while its BLOCK decisions are referred to Sonnet.
At $\omega=0.10$, the Gaussian model selects selective verification
in five cells, flag-to-review cascades in three, and Ministral alone
in one. Its total cost is lower than the Independence selection in
four cells, higher in one, and equal in four.
 
At $\omega=0.50$, the models differ in all nine cells.
Gaussian-based selection has lower evaluation total cost in five
cells and higher total cost in four. For example, at
$(s_L,s_H)=(2,0.5)$, both selected cascades have zero misses,
but the Gaussian selection has 30 false blocks rather than five;
its lower operating cost is offset by the larger false-block loss.
These comparisons show that accounting for dependence can change
the preferred architecture without guaranteeing lower realized cost
in every operating setting.
 
The additional routing flexibility also improves on the optimized
baseline in selected cost--mixture settings. At
$(\omega,s_L,s_H)=(0.10,0.5,2)$, the Gaussian-selected verification
policy invokes Qwen and Ministral, allows when neither blocks,
blocks when both block, and refers disagreement to Sonnet.
It has 16 evaluation misses and seven false blocks, satisfying
the specified evaluation limits, and reduces total cost from
$0.7868$ for the optimized baseline to $0.6010$, a $23.6\%$ decrease.
Across all 37 settings, the Gaussian-selected full-class policy
differs from its baseline selection in eight settings, with lower
evaluation total cost in five and higher cost in three.
 
\paragraph{False-block consequences.}
At $\ell_B=1$, the Gaussian and Independence models select
$\mathrm{Mi}\to H$ and $\mathrm{Mi}\to Q\to G$, with evaluation
total costs $0.7551$ and $1.3989$, respectively.
At $\ell_B=100$, the Gaussian model selects $\pi_{\mathrm V}$,
with zero misses, three false blocks, operating cost $1.0070$,
and total cost $1.2579$ (see the bottom row of Figure~S1 in the Supplementary Material). The Independence model selects
$\pi_{\mathrm I}$, with 14 misses, nine false blocks, operating
cost $0.3779$, and total cost $2.3010$.
Relative to the Gaussian model's optimized baseline $G\to H$,
flexible verification reduces false blocks from five to three,
increases operating cost from $0.8705$ to $1.0070$, and reduces
total cost from $1.2886$ to $1.2579$.
Thus, routing flexibility can improve the balance between service disruption
and invocation cost when false blocks carry high
consequences.
 
Across the 37 settings, the Gaussian model yields lower or equal evaluation
total cost in 32 settings, and higher cost in five.
These results show that dependence can   change risk
assessment and policy selection; realized benefits depend on the
operating requirements and the security--service cost tradeoff.

\section{Discussion and conclusions}
\label{sec:discussion}

This paper develops a dependence-aware framework for assessing
risk and optimizing the design of multi-component
AI safeguard systems. We consider two practical architectures:
sequential layered screening and selective verification.
By modeling the  dependence among safeguard
decisions, the proposed framework accounts for common security
misses and false blocks that cannot be adequately characterized
by individual safeguard performance alone. We further develop
two exact optimization algorithms for selecting safeguard subsets,
execution orders, reviewers, and verification policies under
specified risk and cost requirements. To facilitate practical implementation, we develop DAPOA, an integrated
statistical modeling and optimization package, which is publicly available.

Our theoretical analysis demonstrates that dependence can
substantially affect system risk and operating costs.
For sequential layered screening, the lower-orthant dependence
order provides stochastic comparisons of security misses,
false blocks, and safeguard invocation costs. For selective
verification, a policy's error probabilities respond
monotonically to every supermodular increase in dependence if and
only if its final-decision rule is supermodular or submodular.
Flag-to-review cascades with two or more
first-stage safeguards are not among them: their referral
probabilities and operating costs cannot increase under
supermodular ordering, but the direction of their error
probabilities is undetermined. For a fixed first stage and
reviewer, the optimal routing map is a two-threshold rule in the
first-stage likelihood ratio, and a reviewer is worth invoking only
when its cost is below certain threshold. The numerical experiments further
demonstrate the consequences of ignoring dependence. In six of
the eight dependence scenarios, policies selected under the
independence model violate the specified security-miss limit
when evaluated under the dependent distribution. These findings
show why joint safeguard decisions must be considered when
assessing risk and selecting a safeguard policy.

The empirical study provides further evidence using actual LLM
safeguard decisions. It evaluates the policy choices made using
both fitted class distributions on held-out actions. Under the principal operating setting,
the Gaussian-based selection records zero security misses among
598 unauthorized actions, compared with 14 for independence-based
selection, while false blocks increase from nine to 30 among
598 authorized actions. Operating cost increases from $0.3779$
to $0.7300$, and evaluation total cost decreases from $1.6237$
to $0.9809$, a reduction of $39.6\%$. Across the 37 operating
settings, dependence modeling changes the selected policy in
22 settings and yields lower evaluation total cost in 17 settings,
higher cost in five, and equal cost in 15. Selective verification
provides additional gains over the optimized baseline in some
operating settings. For example, at
$(\omega,s_L,s_H)=(0.10,0.5,2)$, flexible routing reduces
evaluation total cost from $0.7868$ to $0.6010$, a $23.6\%$
decrease. These results demonstrate how dependence information
and routing flexibility affect the security service cost tradeoff.

Several limitations warrant consideration. First, the empirical
study evaluates actual LLM safeguard decisions on benchmark-derived
agent actions rather than observations from live deployments.
The controlled setting provides well-defined authorization labels
and complete safeguard decisions, allowing alternative policies
to be evaluated consistently on identical actions.
Further validation under production operating conditions would
help establish the generalizability of these findings. Second, invocation costs are normalized from observed token usage
under specified pricing assumptions, while the consequences of
security misses and false blocks are represented by assumed loss
parameters. Obtaining reliable estimates of these costs is
challenging, as they depend on the operating environment and
the potentially substantial downstream consequences of safeguard
errors. Nevertheless, the proposed framework can directly
incorporate deployment-specific cost estimates when such
information becomes available. Third, the current framework assumes that the dependence structure remains fixed during policy optimization.
Extending the methodology to accommodate time-varying dependence
and adaptive attacks would be valuable, particularly when
safeguard performance changes with the operating environment.

Overall, our findings highlight the importance of dependence
modeling in the risk assessment and optimal design of
AI safeguard systems. By integrating statistical dependence
modeling with risk-constrained optimization, this work
provides a foundation for systematic safeguard design beyond
individual component performance. The operational decisions include
which safeguards to include in a policy, their invocation order,
and which first-stage outcomes justify additional review. Further
work can extend this formulation to larger candidate pools, changing
operating mixtures, and adaptive safeguard policies.

\FloatBarrier

\bibliographystyle{apalike-ejor-nolinks}
\bibliography{references_ejor}

\end{document}